\documentclass[lettersize,journal]{IEEEtran}
\usepackage{amsmath,amsfonts}
\usepackage{algorithmic}
\usepackage{algorithm}
\usepackage{array}
\usepackage[caption=false,font=normalsize,labelfont=sf,textfont=sf, subrefformat=parens]{subfig}
\usepackage{textcomp}
\usepackage{stfloats}
\usepackage{url}
\usepackage{verbatim}
\usepackage{graphicx}
\usepackage{hyperref}
\usepackage{cite}
\floatname{algorithm}{Process}

\begin{document}

\title{SVDE: Scalable Value-Decomposition Exploration for Cooperative Multi-Agent Reinforcement Learning}

\author{Shuhan Qi, Shuhao Zhang, Qiang Wang, Jiajia Zhang, Jing Xiao, Xuan Wang
\thanks{Corresponding author: Qiang Wang, Jiajia Zhang.}
\thanks{Shuhan Qi is with the School of Computer Science and Technology, Harbin Institute of Technology, Shenzhen, Guangdong 518055, and also with the PengCheng Laboratory, Shenzhen, Guangdong 518055, China (e-mail: shuhanqi@cs.hitsz.edu.cn).}%
\thanks{Shuhao Zhang is with the School of Computer Science and Technology, Harbin Institute of Technology, Shenzhen, Guangdong 518055, China.}%
\thanks{Qiang Wang is with the School of Computer Science and Technology, Harbin Institute of Technology, Shenzhen, Guangdong 518055, China.}%
\thanks{Jing Xiao is with the Ping An Technology(Shenzhen) Artificial Intelligence and Big Data Center, No. 20, Keji South 12th Rd, Nanshan, Shenzhen, Guangdong 518055, China.}%
\thanks{Jiajia Zhang and Xuan Wang are with the School of Computer Science and Technology, Harbin Institute of Technology, Shenzhen, Guangdong 518055, and also with the Guangdong Provincial Key Laboratory of Novel Security Intelligence Technologies, Shenzhen, Guangdong 518055, China, (email: zhangjiajia@hit.edu.cn).}%
}



\maketitle

\begin{abstract}
Value-decomposition methods, which reduce the difficulty of a multi-agent system by decomposing the joint state-action space into local observation-action spaces, have become popular in cooperative multi-agent reinforcement learning (MARL).
However, value-decomposition methods still have the problems of tremendous sample consumption for training and lack of active exploration.
In this paper, we propose a scalable value-decomposition exploration (SVDE) method, which includes a scalable training mechanism, intrinsic reward design, and explorative experience replay.
The scalable training mechanism asynchronously decouples strategy learning with environmental interaction, so as to accelerate sample generation in a MapReduce manner. 
For the problem of lack of exploration, an intrinsic reward design and explorative experience replay are proposed, so as to enhance exploration to produce diverse samples and filter non-novel samples, respectively. 
Empirically, our method achieves the best performance on almost all maps compared to other popular algorithms in a set of StarCraft II micromanagement games.
A data-efficiency experiment also shows the acceleration of SVDE for sample collection and policy convergence, and we demonstrate the effectiveness of factors in SVDE through a set of ablation experiments.
\end{abstract}


\section{Introduction}
\IEEEPARstart{C}{ooperative} multi-agent reinforcement learning (MARL) \cite{tnnls_marl,tnnls_cooperative_task} has made great progress such as in solving sparse rewards \cite{sparse,sparse2}, incorporating game theory and agent communication \cite{game1,coma,communication}, and combining relationships between agents \cite{relation}.
However, there are two unsolvable problems in MARL.
First is  the \emph{data-efficiency} problem  under the framework of centralized training with decentralized execution (CTDE) \cite{tnnls_dataefficiency}, which is mainly caused by poor sample generation and an unbalanced sample distribution in the replay buffer.
The learning of MARL requires large samples, which is time-consuming due to the high cost of interacting with the environment.
The replay buffer of off-policy reinforcement learning (RL) is always stuffed by trajectories of various qualities, the high quality sample can hardly be selected for training, which makes convergence slower.
There is also the \emph{exploration and exploitation} problem \cite{tnnls_exploration1,tnnls_exploration2,tnnls_exploration3} in MARL.
As the number of agents scales up, the dimensionality of the joint state-action space grows exponentially, which   makes exploration more difficult\cite{marlexploration}. 
As a result, the agent will easily fall into local optima.

\begin{figure}[t]
    \begin{minipage}[t]{1\linewidth}
    \centering
    \subfloat[]{
    \includegraphics[width=3.8cm]{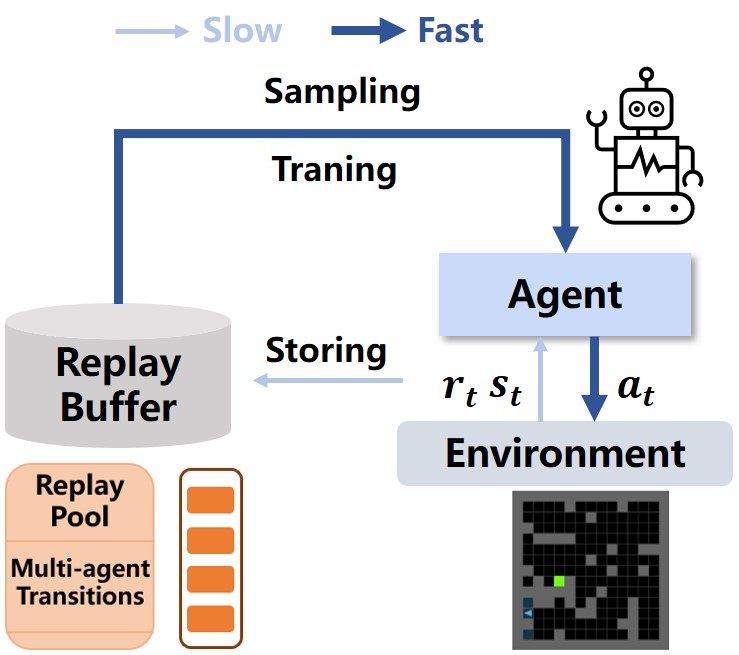}
    }\noindent
    \subfloat[]{
    \includegraphics[width=3.8cm]{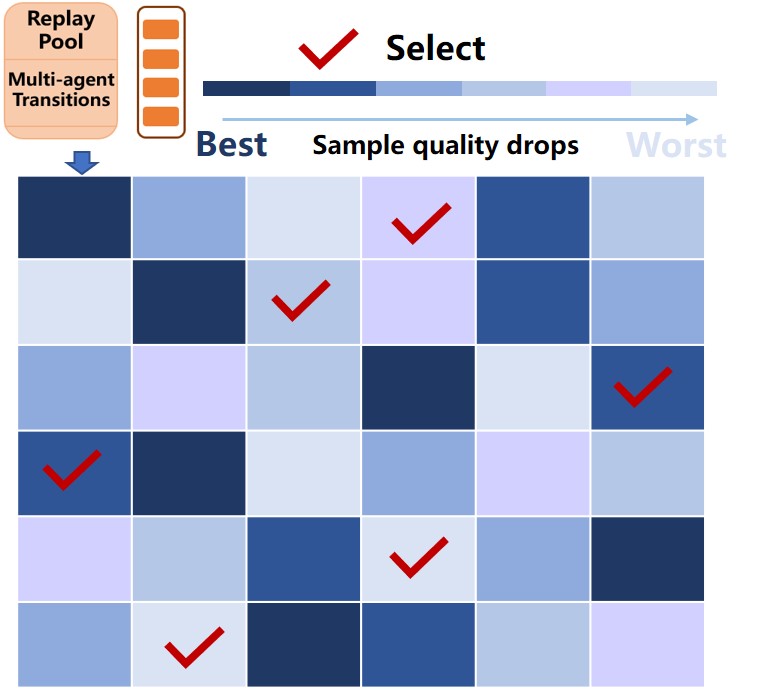}
    }
    \end{minipage}
    \begin{minipage}[t]{1\linewidth}
    \centering
    \subfloat[]{
    \includegraphics[width=8cm]{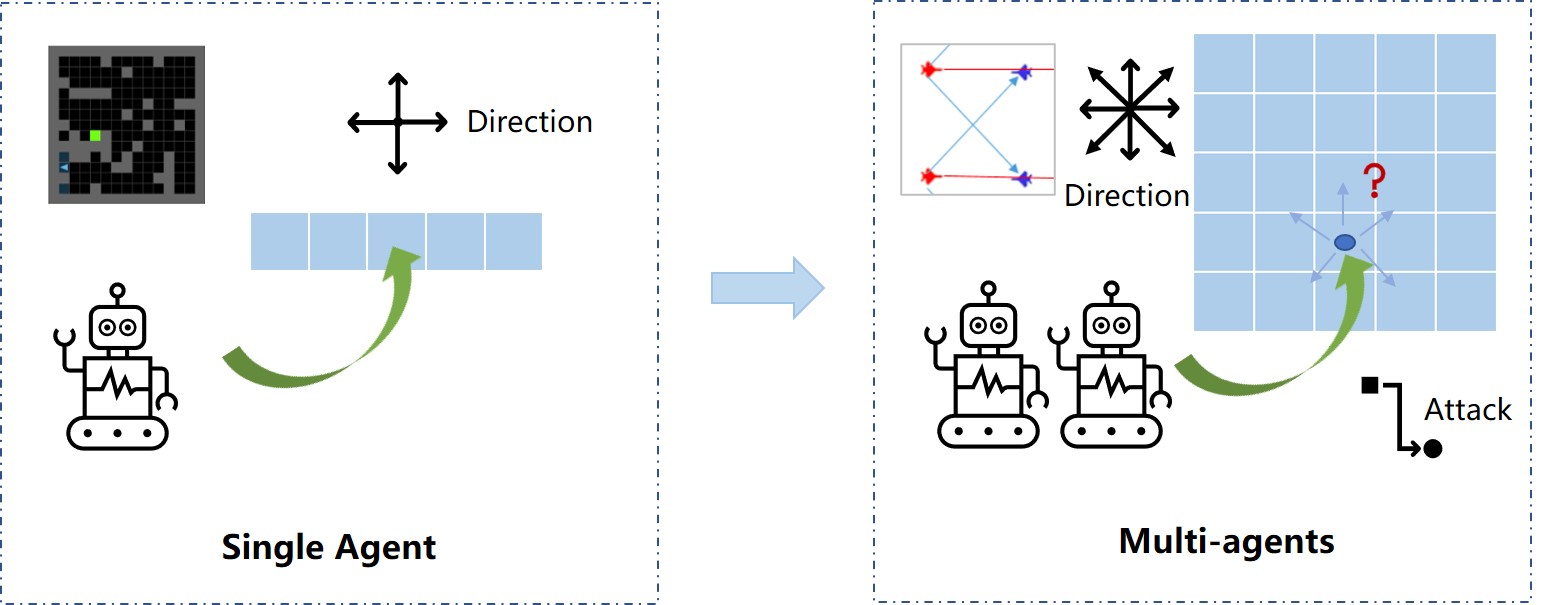}
    }
    \end{minipage}
\caption{Two problems. \textbf{\emph{data-efficiency}}: $(a)$ with poor sample generation, insufficient samples lead to reduced sample diversity and quality; $(b)$ unbalanced sample distribution in   replay buffer: uniform sampling of existing methods leads to better samples not being selected. \textbf{\emph{exploration and exploitation}}: $(c)$ exploration becomes more difficult as   joint state-action space expands exponentially.}
\label{fig:1}
\vskip -0.3cm 
\end{figure}

There are many efficient exploration methods in single-agent reinforcement learning (SARL), such as environment reset and recording action paths in an archive \cite{go-explore}.
Due to the exponential growth of the state-action space, these methods cannot be simply migrated to MARL \cite{tnnls_sa_to_ma}.
Distributed training frameworks such as IMPALA \cite{impala} and SEED \cite{seed} only support SARL and are unsuitable for MARL.
Although Ray \cite{ray} supports some specific MARL algorithms, it does not help the agent better explore the environment and efficiently use sample data.


Value-decomposition is a popular method in MARL.
By decomposing the joint state-action space into local observation-action spaces, value-decomposition can significantly reduce the difficulty of multi-agent learning. For example, 
VDN \cite{vdn} decomposes the joint action-value function into the sum of multiple independent agent action-value functions.
QMIX \cite{qmix} and WQMIX \cite{wqmix} relax the assumption to a nonlinear monotonic combination of local action-value functions, and are most related to our method. 
But they can only perform passive exploration (such as $\epsilon-$greedy), and cannot effectively and actively explore the environment.
Some recent work like QTRAN \cite{qtran} and MAVEN \cite{maven} follows the way of value-decomposition and further improves the ability of multi-agent systems.
However, these methods still cannot solve the above problems.

We propose a scalable value-decomposition exploration (SVDE) algorithm that has a scalable training mechanism, intrinsic reward design, and explorative experience replay.

First, the scalable training mechanism in SVDE adapts to cooperative MARL, which collects a large number of samples in a MapReduce manner \cite{mapreduce}.
It models the general MARL learning problem as three processes: \emph{rollout}, \emph{serving}, and \emph{training}, and contains four main modules: \emph{replay buffer}, \emph{actor}, \emph{worker}, and \emph{a centralized learner}.
The scalability of workers and actors can make full use of computing resources, and such a function greatly expands the parallelization of sample generation.
Through the three processes and four modules, learning and training can be decoupled, which accelerates the convergence of the joint Q-function.

Second, the intrinsic reward design enables the agent to perform effective exploration when the environment only returns a team reward, i.e., an individual agent does not receive its own reward.
It worth noting that we combine the intrinsic reward of curiosity with the value-decomposition method. 
Curiosity rewards encourage agents to generate diversified trajectories, which makes for better exploration behavior \cite{liir}. 
Based on the value-decomposition method, each agent can calculate intrinsic rewards only based on its local observations, without considering the influence of the exploration behavior on the joint action of multi-agents.

Third, SVDE uses explorative experience replay to make better use of training samples. 
When selecting samples from the replay buffer, different from other priority experience replay (PER) \cite{rnd}, explorative experience replay considers the “freshness” of samples as well as the priority. 
We adopt an Upper Confidence Tree (UCT)-like method to filter and replace the outdated samples from the replay buffer in time, so as to further force the model to keep exploring all the time.

In multiple maps of StarCraft II, SVDE has achieved the best experimental results compared with popular MARL algorithms, with an average win rate exceeding 90\% on all maps.
And the results from data-efficiency and sample distribution experiments fully demonstrate the acceleration of the scalable training mechanism for sample collection and policy convergence.
Moreover, our ablation experiments show both the necessity of intrinsic reward design and explorative experience replay in order to explore comprehensively and learn better strategies.

\section{Related Work}
Recent work in MARL has shown a diversified development with the progress of deep learning, and has gradually moved from the tabular methods \cite{tabular} to various deep neural networks \cite{deeplearning}.
Our method is based on the cooperative value-decomposition algorithm, which is related to the recent advances in CTDE \cite{tnnls_dataefficiency}.
Specifically, our method takes advantage of the efficiency and scalability of CTDE to improve the performance of cooperative MARL in large-scale environments.

In MARL, one challenge is to learn effective policies in a decentralized environment. 
Prior works have explored different approaches to this problem. 
On the one hand, some methods directly learn decentralized value functions or policies, which are simple and intuitive.
Independent Q-learning (IQL) \cite{iqlta, iql} integrates q-learning into multi-agent systems, learning a separate value function for each agent and enabling fully decentralized learning. 
However, these methods can suffer from environmental non-stationary problems during the training process, induced by the policy update and exploration of each agent simultaneously. 
Stabilising experience replay \cite{stabilising} addresses learning stabilization under the decentralized paradigm, but still cannot completely solve this problem.
On the other hand, centralized methods can handle non-stationary problems at the cost of scalability. 
Coordinated reinforcement learning \cite{center1} exploits the conditional independence between agents to decompose the global reward into the sum of each agent's local rewards.
Sparse cooperative Q-learning \cite{center2} uses state sparsity to tabulate the state, allowing the use of tabular Q-learning.
These methods have shown promising results in various applications, but may have limitations in scalability or complexity.

A common method is to balance the benefits of centralized and decentralized learning
One compromise is to use global information for centralized training and decentralized execution.
COMA \cite{coma} is a policy-based MARL algorithm in Counterfactual Multi-Agent Decision Making (CMDM) to estimate the counterfactual baseline through a centralized critic.
Similarly, MARS \cite{samecoma} is easier to scale to more agents but reduces the centralization advantage.
However, these methods use on-policy policy gradient learning, which can be data-inefficient and prone to suboptimal results.
Existing distributed reinforcement learning (RL) frameworks such as IMPALA \cite{impala} and SEED \cite{seed} only support SARL algorithms, while Ray \cite{ray}, TLeague \cite{tleague}, and MAVA \cite{mava} not only have high requirements for computing resources but are not fully adapted to cooperative MARL.

Recently, value-based methods that are intermediate between IQL and COMA have achieved great success in complex multi-agent tasks \cite{tnnls_cooperative_task2}. 
These methods decompose the joint action-value function into functions of independent action-value functions by satisfying the constraints of Individual-Global-Max (IGM) consistency \cite{qtran}. 
VDN \cite{vdn}, QMIX \cite{qmix}, and WQMIX \cite{wqmix} enforce additivity, monotonicity, and weighted monotonicity constraints, respectively. 
Qatten \cite{qatten} uses both linearity and monotonicity constraints.
QTRAN \cite{qtran} and Qplex \cite{qplex} completely realize the expression ability of value-decomposition by converting IGM to optimization constraints and by duplex dueling architecture, respectively, but they are found to have high computational complexity in experiments.
However, none of the above methods can help agents explore high-dimensional multi-agent systems efficiently, which limits their performance.

While some works has been made in multi-agent reinforcement learning (MARL) exploration, it is still in its early stages due to the difficulty of exploring the joint state-action space.
Several methods have been proposed to address this issue.
EUIR \cite{exploration1} introduces a centralized agent with intrinsic rewards to interact with the environment, and stores experience in a shared replay buffer to update policies. 
CEIR \cite{exploration2} defines several types of intrinsic rewards by combining the decentralized curiosity of each agent.
LIIR \cite{liir} learns an additional proxy critic for each agent to learn the individual intrinsic reward and uses it to update the policy.
However, these types of intrinsic rewards are domain-specific and cannot be extended to other scenarios.
MAVEN \cite{maven} leverages a hierarchical policy to control shared latent variables as a signal for coordinated exploration patterns, but it still has limitations in some scenarios such as reward sparse environments.

In this paper, we study how to solve the problems of low data-efficiency and insufficient exploration in the value-decomposition method.

\section{Preliminaries}
\subsection{Model and CTDE}
\subsubsection{Dec-POMDP}
We consider a fully cooperative multi-agent task as a decentralized partially observable MDP \cite{pomdp} consisting of a tuple $\mathcal{G}=(\mathbf{n}, \mathcal{S}, \mathcal{A}, \mathcal{P}, \mathcal{R}, \mathcal{Z}, \mathcal{O}, \gamma)$ in which $n$ represents the number of agents in the environment and $|\mathbf{n}|=n$.
$s \in \mathcal{S}$ describes the true state of the environment.
$\mathcal{A}$ represents the set of actions that the agent can take, and at each time step, each agent $i \in\{1, \ldots, n\}$ chooses action $a^i \in \mathcal{A}$, forming a joint action $\mathbf{a} \in \mathbf{A} \equiv \mathcal{A}^n$.
$\mathcal{P}$ is the state transition function of the environment.
Function $P\left(s^{\prime} \mid s, \mathbf{a}\right): \mathcal{S} \times \mathbf{A} \times \mathcal{S} \rightarrow[0,1]$ represents the probability that the environment transitions from the current state $s$ to a new state $s^{\prime}$ after taking a joint action $\mathbf{a}$.
All agents share the same global reward function $r(s,\mathbf{a}): \mathcal{S} \times \mathbf{A} \rightarrow \mathbb{R}$ and $\gamma \in [0,1)$, where the $\gamma$ is a discount factor for controlling reward weights at different time steps.
Furthermore, each agent only draws a partial observation $z \in \mathcal{Z}$ according to the observation function $\mathcal{O}(s,a^i): \mathcal{S} \times \mathcal{A} \rightarrow \mathcal{Z}$ and learns an individual stochastic-policy $\pi^i\left(a^i \mid \tau^i\right): \mathcal{T} \times \mathbf{A} \rightarrow [0,1]$ conditioned on its local observation-action history $\tau^i \in \mathcal{T} \equiv (\mathcal{Z} \times \mathbf{A})^*$.
The objective of all agents is to maximize the \emph{cumulative discount return}$: \mathcal{R}_t=\sum_{l=0}^{\infty} \gamma^i r_{t+l}$.
The joint action-value function produced under the joint policy $\pi$:
$$
Q^\pi\left(s_t, \mathbf{a}_t\right)=\mathbb{E}_{s_{t+1 ; \infty}, \mathbf{a}_{t+1 ; \infty}}\left[\mathcal{R}_t \mid s_t, \mathbf{a}_t\right]
$$

\subsubsection{CTDE}
Most early methods in MARL tasks directly learn an independent action-value function for each agent, i.e., IQL \cite{iql}.
Although this method is scalable and intuitive, it causes a non-stationary environment during training and can be difficult to converge.
Recently, centralized training with decentralized execution (CTDE) has emerged as an alternative solution.
This method allows agents to obtain all local observation-action histories $\mathcal{\tau}$ and the global state $s$.
However, each agent's local policy still relies on its own observation-action history $\boldsymbol{\tau}^i$ when making decisions.

\subsection{Value-decomposition and IGM Condition}
Value-decomposition \cite{vdn,qmix,wqmix,qtran,qatten,qplex} is a solution used to simplify the process of finding optimal joint Q-function in complex joint state-action space.
An important concept in this method is Individual-Global-Max (IGM) \cite{qtran}, which ensures the consistency of the combination of the local individual optimal actions of each agent and the global optimal action.
The formula expression of IGM is shown in 

\begin{equation}
\arg \max _{\mathbf{a}} Q^{\mathrm{jt}}(\boldsymbol{\tau}, \mathbf{a})=\left(\begin{array}{c}
\arg \max _{a^1} Q^1\left(\tau^1, a^1\right) \\
\vdots \\
\arg \max _{a^N} Q^n\left(\tau^n, a^n\right) 
\end{array}\right) 
\label{equ:1}
\end{equation}

VDN \cite{vdn} and QMIX \cite{qmix} propose the \emph{additivity} and \emph{monotonicity} assumptions to satisfy the IGM condition, respectively.

\begin{equation}
(\text { VDN }) \quad  Q^{\mathrm{jt}}(\boldsymbol{\tau}, \mathbf{a})=\sum_{i=1}^\mathbf{n} Q^i\left(\tau^i, a^i\right)
\label{equ:2}
\end{equation}

\begin{equation}
(\text { QMIX }) \quad \frac{\partial Q^{\mathrm{jt}}(\boldsymbol{\tau}, \mathbf{a})}{\partial Q^i\left(\tau^i, a^i\right)}>0, \forall i \in \mathbf{n}
\label{equ:3}
\end{equation}

\subsection{Distributed RL and Exploration}
\subsubsection{Distributed RL}
Distributed RL aims at arranging multiple actors to interact with environments synchronously and collect training samples for centralized training, thereby accelerating the convergence speed of training process.
Researchers have proposed multiple distributed RL frameworks, among which IMPALA \cite{impala} and SEED \cite{seed} have achieved state-of-the-art (SOTA) results on benchmarks.

\begin{figure}[h]
    \subfloat[]{
    \begin{minipage}[t]{0.5\linewidth}
    \includegraphics[width=4cm]{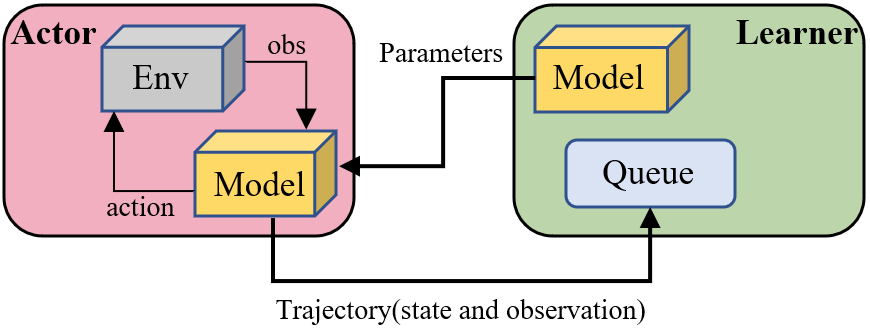}
    \label{subfig:1}
    \end{minipage}
    }
    \subfloat[]{
    \begin{minipage}[t]{0.5\linewidth}
    \includegraphics[width=4cm]{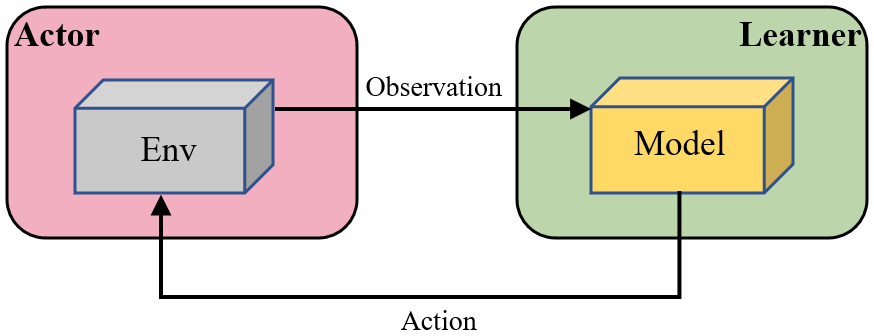}
    \label{subfig:2}
    \end{minipage}%
    }
\caption{(a) IMPALA and (b) SEED.}
\label{fig:3}
\end{figure}

Fig.~\ref{fig:3} shows the training process of IMPALA and SEED.
The actor module in IMPALA contains a forward network for decision-making that periodically synchronizes parameters from the learner.
The actor transfers the data to the learner for the training of the central network after collecting a certain amount of samples.
In contrast, the actor module in SEED is responsible only for interacting with the environment, and decision-making is the responsibility of the network of the learner. 
The efficiency of SEED is higher than that of IMPALA because the actor is generally on the CPU unit, and the learner is on the GPU unit. 
Moreover, the communication cost of transferring network parameters and data in IMPALA is too high, which can also cause communication bottlenecks.

\begin{figure*}[htbp]
    \centering
    \includegraphics[height=8cm, width=14cm]{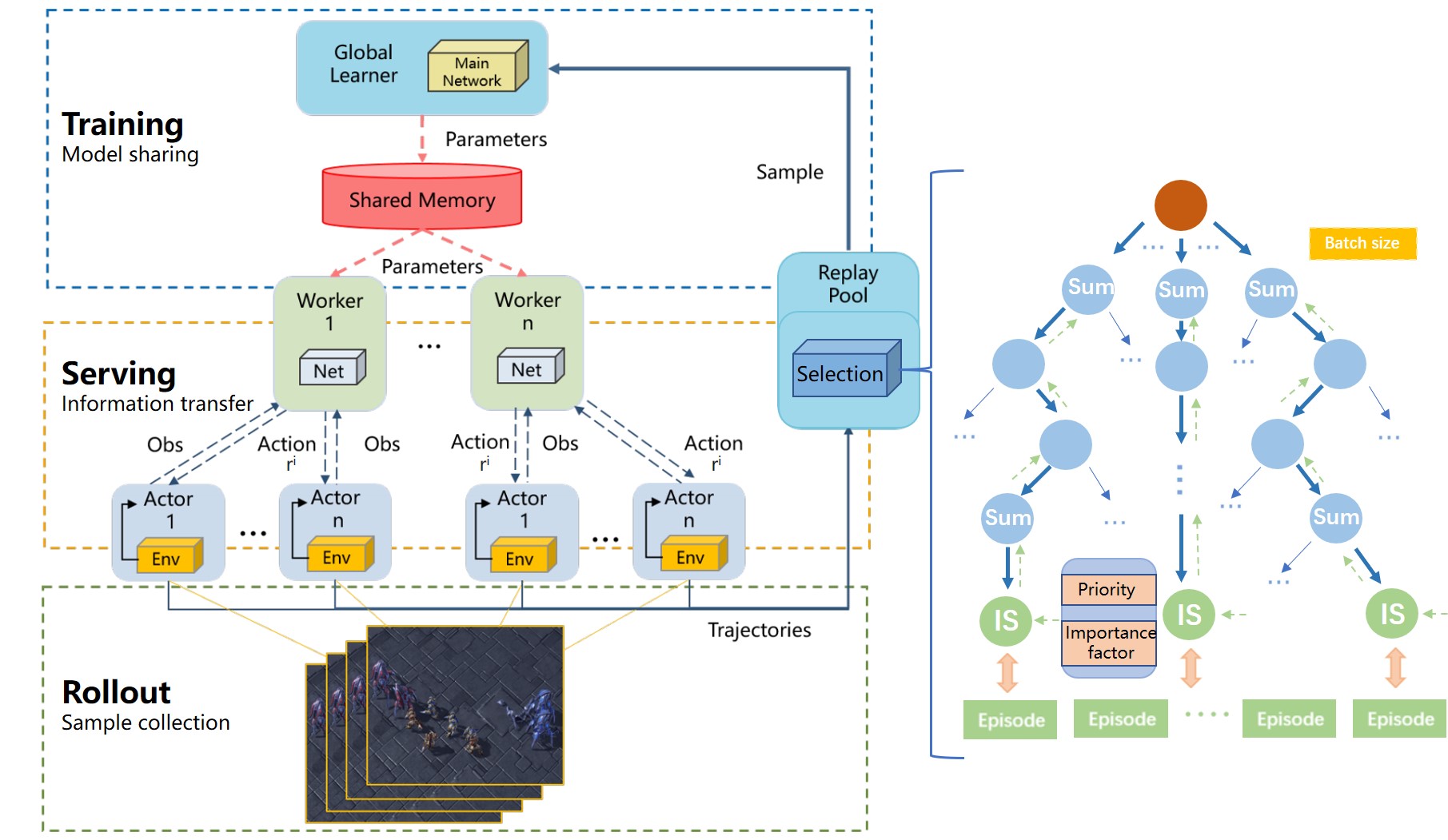}
    \caption{Efficient scalable training mechanism. Three-layer structure corresponds to Learner, Worker, Actor, who cooperate from top to bottom to complete   processes of training, serving, and rollout. Replay buffer on   right contains a batch selection module (see section.~\ref{eeif}).}
    \label{fig:4}
\end{figure*}

\subsubsection{Intrinsic Motivation and UCB}
Upper confidence bounds (UCB) is a classic exploration method that has been used in various applications, including AlphaGo \cite{alphago} which also uses its variant upper confidence tree (UCT) to explore.
Compared with traditional random exploration methods that are often inefficient, the methods in the UCB family measure the potential of each action by an upper confidence bound of the expected reward.
Each selection of actions can be calculated as

\begin{equation}
a=\arg \max _a Q(a)+\sqrt{\frac{-\ln k}{2 \mathcal{N}(a)}} ,
\label{equ:4}
\end{equation}
where $Q(a)$ is the expected reward of action $a$, $\mathcal{N}(a)$ is the count to select action $a$ and $k$ is a constant hyperparameter.

Intrinsic motivation-oriented exploration methods are primarily designed to create appropriate intrinsic rewards to assist the agent in exploring the environment.
These methods are particularly useful when the environment returns sparse external rewards or only a team reward in MARL.
Prediction Error is a form of intrinsic reward design that has shown promising results on many SARL tasks \cite{rnd,icm}.
Formally, it uses the prediction error of the next state as an intrinsic reward, which can be measured by the difference between the predicted and the real next state:

\begin{equation}
R\left(s_t, s_{t+1}\right)=\left\|g\left(s_{t+1}\right)-\hat{F}\left(g\left(s_t\right), a_t\right)\right\|_2 ,
\label{equ:5}
\end{equation}
where $g$ is an encoder that encodes the state into a low-dimensional feature and $\hat{F}$ models the environmental dynamics.
How to properly set $\hat{F}$ and learn a suitable $g$ is the main challenge.

\section{Methodology}
The proposed SVDE method follows the following constraints and assumptions:
$\text { (1) }$ it follows the same monotonicity assumption as equation \eqref{equ:3};
$\text { (2) }$ the global state is only used in training, and the learned Q-function only relies on each agent's own local observation-action history to make decisions, using no other external information;
$\text { (3) }$ parameters of agents are shared, i.e., we train only one network for all the agents;
$\text { (4) }$ there is no form of communication between agents during training;
$\text { (5) }$ it is a value-based off-policy cooperative MARL method;
$\text { (6) }$ the environment's dynamic model is unaware, i.e., it is  model-free.


SVDE has three parts.
\textbf{The Efficient Scalable Training mechanism} collects a large number of samples by asynchronously arranging multiple \emph{actors} and \emph{workers};
\textbf{Intrinsic Reward Design} encourages exploration through intrinsic rewards for states to solve the poor quality of generated samples;
\textbf{Explorative Experience Replay} optimally selects samples in the replay buffer for training.
 
\subsection{Efficient Scalable Training Mechanism}
We propose an efficient scalable training mechanism for cooperative MARL methods, which has four modules.

\textbf{Actor.} Each actor is responsible for interacting with the environment in which it is located, and the environments in which different actors interact are completely independent.
Decision-making is the responsibility of the worker, so as to avoid the communication bottleneck caused by transferring parameters and trajectories between actors and workers, such as in IMPALA \cite{impala}.

\textbf{Worker.} Each worker is responsible for the decision-making and intrinsic reward calculation of multiple actors.
The agent network of the worker consists of an action network and an IRD network (described below), which are responsible for making decisions and calculating intrinsic rewards, respectively.
The agent network periodically synchronizes parameters from the shared memory, i.e., it does not participate in training.

\textbf{Replay Buffer.} After the actor completes a round of interaction with the environment, the trajectory is stored in the replay buffer for the learner's training.
A selection module in the replay buffer selects training data based on the priority and importance factor (described below) of samples instead of uniform sampling.

\textbf{Learner.} The network of the Learner consists of a double mixing network and multiple agent networks. 
The learner continuously samples batch data from the replay buffer for training, and periodically synchronizes the latest parameters of the agent network to the network of the worker through the shared memory.

We consider distributed RL in the following three processes, with the architecture shown in Fig.~\ref{fig:4}.
\begin{itemize}
    \item \textbf{Rollout.} The actor interacts with the environment and collects samples to the replay buffer;
    \item \textbf{Serving.} The worker transfers the action and intrinsic reward to the actor through \emph{information transfer} after finishing calculations;
    \item \textbf{Training.} The learner trains the network and regularly updates the parameters of the agent network to the worker through \emph{model sharing}.
\end{itemize}


\subsubsection{Rollout}
The rollout process is shown as Process \ref{pro:1}.
When the actor receives a new observation of the environment, it transfers it into the observation-action pipeline and  retrieves the action and intrinsic reward from the same pipeline. 
It is similar to SEED \cite{seed} and MAVA \cite{mava}, but they do not take into account the exploration aspects.
The actor holds the entire observation-action history, i.e., the trajectory, and transfers it to the replay buffer for the learner's training at the end of a game.
\begin{algorithm}[h]
\caption{Rollout} 
\textbf{Input:} environment parameters $args$,
                              interaction times $T$\\ 
\textbf{Global:} sample queue $Q$,
                            observation-action pipeline $P$
                         
\begin{algorithmic}[1]
\STATE \textbf{Function} Actor($args$, $T$)\textbf{:}
    \STATE Construct environment $Env$ from environment parameters $args$;
    \STATE $Env.start()$;\\
    \STATE Initialize interaction times $t \leftarrow 0$;\\
    \REPEAT
        \STATE $Env.reset()$;
        \STATE Get the start state;
        \REPEAT
            \STATE Put the observation $o$ into $P$;
            \STATE Get action $a$ from $P$;
            \STATE $o,r \leftarrow Env.step(a)$;
        \UNTIL{$Env.done()$};
        \STATE $t \leftarrow t+1$;
        \STATE Send the collected trajectory to $Q$;
    \UNTIL{$t > T$};
    \STATE $Env.close()$;
\RETURN
\end{algorithmic}
\label{pro:1}
\end{algorithm}

\subsubsection{Serving}
The worker deploys multiple actors, gets observations from the observation-action pipeline, makes decisions, and calculates intrinsic rewards using the agent network (described below).
It returns actions and intrinsic rewards to the actors via the observation-action pipeline.
The agent network periodically synchronizes parameters from the network of the learner.
Therefore, network training and environmental interaction are decoupled, and it is ensured that the agent network can obtain the latest network parameters in real time for serving.
In addition, since the network is uniformly trained by the learner, it is possible to use the samples collected by all workers to train the network, increasing its robustness, instead of just using the samples collected by the current worker.
When the worker fails to get new observations from the observation-action pipeline, the loop ends.
\begin{algorithm}[h]
\caption{Serving} 
\textbf{Input:}   number of actors $n$,
                              total interaction times $nT$\\ 
\textbf{Global:} network parameters $\theta$,
                            shared memory $M$,
                            observation-action pipeline $P$
                         
\begin{algorithmic}[1]
\STATE \textbf{Function} Worker($n$, $nT$)\textbf{:}
    \STATE Deploy $n$ actors and perform $T$ environment interactions;
    \REPEAT
        \STATE Get parameters $\theta$ from $M$;
        \STATE Get observations $o$ from $P$;
        \STATE Make decisions according to the local Q-function, \\
        $a=\underset{a}{\arg \max } Q^i(o, a)$; \\
        \emph{ (the specific $r^i$ definition will be introduced in subsection~\ref{ird}) }
        \STATE Calculate intrinsic rewards $r^i$ according to the prediction error;
        \STATE Send actions $a$, intrinsic rewards $r^i$ into $P$;
    \UNTIL{$P.size == 0$};
\RETURN
\end{algorithmic}
\label{pro:2}
\end{algorithm}

\begin{algorithm}[h]
\caption{Training} 
\textbf{Input:}   number of workers $m$,
                              batch size $B$\\ 
\textbf{Global:} replay buffer $E$,
                            shared memory $M$,
                            sample queue $Q$,
                            network parameters $\theta$
                         
\begin{algorithmic}[1]
\STATE \textbf{Function} Learner($m$, $B$)\textbf{:}
    \STATE Initialize network $\theta,E,Q,M$;
    \STATE Deploy $n$ workers;
    \REPEAT
        \STATE Get data from $Q$ and store it in $E$;        
        \STATE Randomly sample $B$ samples from $E$;
        \STATE Calculate the gradient:\\
        \emph{($Q^{\mathrm{jt}}$ is defined in subsection~\ref{method} ) }   \\
        $d \theta \leftarrow\left(y_{j}^{\mathrm{jt}}-Q^{\mathrm{jt}}(s, \mathbf{a} ; \theta)\right) \nabla_{\theta} Q^{\mathrm{jt}}(s, \mathbf{a} ; \theta)$;
        \STATE Update parameter $\theta$:\\
        $\theta \leftarrow \theta+d \theta$;
        \STATE Sync parameters $\theta$ to $M$;
    \UNTIL{$Q.size == 0$};
\RETURN
\end{algorithmic}
\label{pro:3}
\end{algorithm}
\subsubsection{Training}
The learner asynchronously deploys multiple workers and continuously samples batch data from the replay buffer through a sample queue.
Training continues while the sample queue is not empty, i.e., training and rollout are asynchronous.
Network parameters in shared memory are updated periodically to facilitate the updating of the agent network of workers.
This ensures that multiple workers asynchronously obtain the latest network parameters to update the local action-value function.
When the sample queue is empty, training ends.

\begin{figure}[htbp]
    \centering
    \includegraphics[width=9cm]{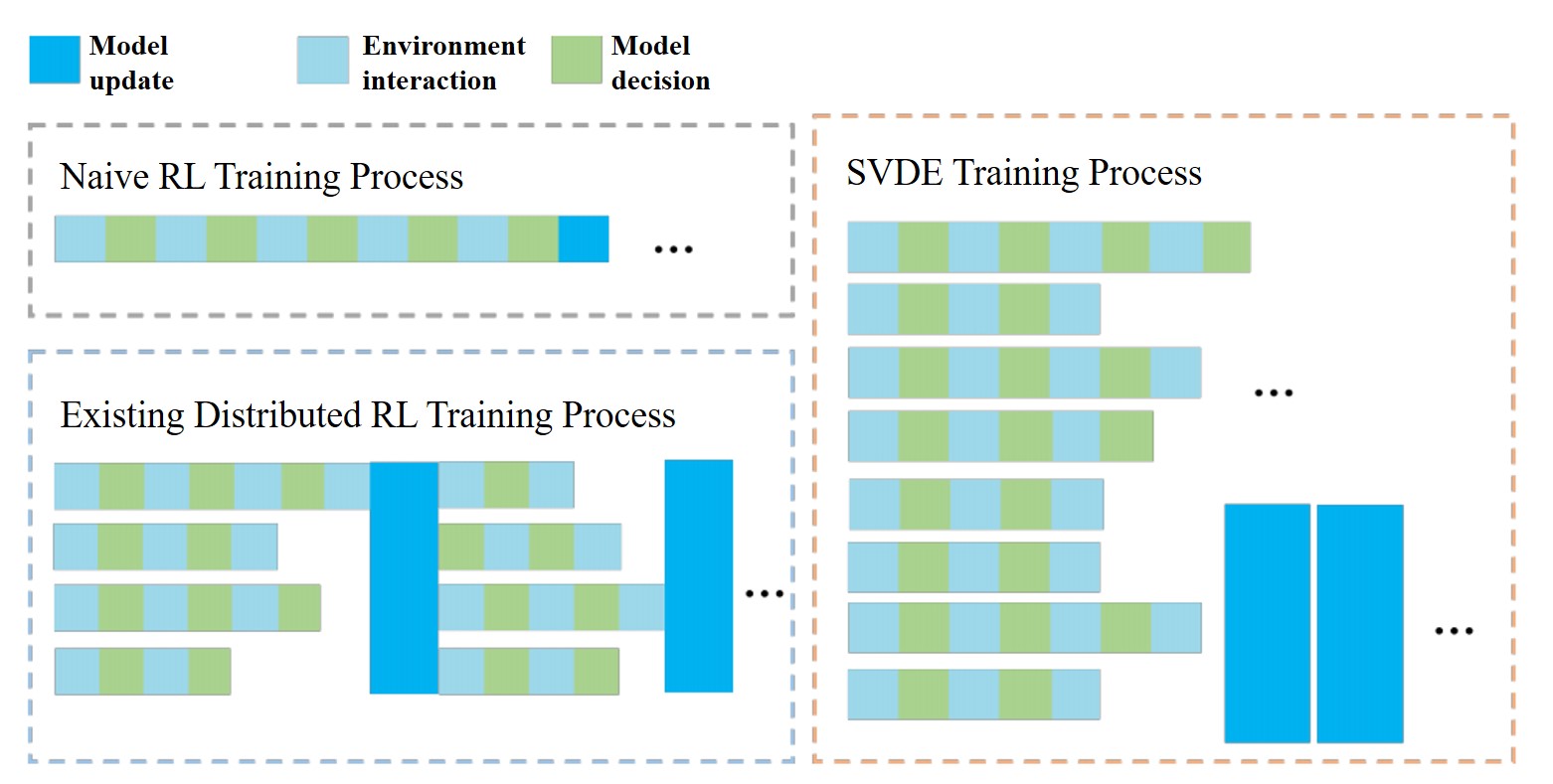}
    \caption{Comparison of training processes.}
    \label{fig:5}
\end{figure}

Through our proposed scalable training mechanism, we decouple the rollout and training processes, as shown in Fig.~\ref{fig:5}.
The number of actors and workers can be configured according to the environment and requirements.
It is possible to use GPU units to accelerate the rollout process, which is usually performed on CPUs in traditional RL training \cite{impala,seed,ray}.

\subsection{Intrinsic Reward Design and Explorative Experience Replay}
\subsubsection{Intrinsic Reward Design (IRD)}
To encourage agents to explore in SARL tasks, prior works generally hope to visit novel states more frequently, i.e., those states visited less often \cite{marlexploration}.
Count-based exploration methods provide an example of such bonuses, where the intrinsic reward is generally $r^i=1/{n(s)}$, where $n$ is the number of visits to the state \cite{countbased}.
As the problem becomes more complex, the state space gradually increases, and almost every state can only be explored once.
An alternative is to calculate the prediction error associated with the agent's transition \cite{rnd}.
Since the environment only feeds back a team reward, there exists a credit assignment problem in MARL tasks.
We introduce an intrinsic reward based on prediction error, and calculate it for each agent's local observations, so as to encourage the agent to explore.
In this way, we extend the intrinsic reward design from SARL \cite{rnd }to MARL \cite{tnnls_sa_to_ma}.
And since each agent calculates its intrinsic reward based on its own local observations, the intrinsic reward can be used to represent the contribution of the agent to the whole, thus solving the problem of credit assignment to a certain extent.

\begin{figure*}[t]
    \centering
    \includegraphics[width=17cm]{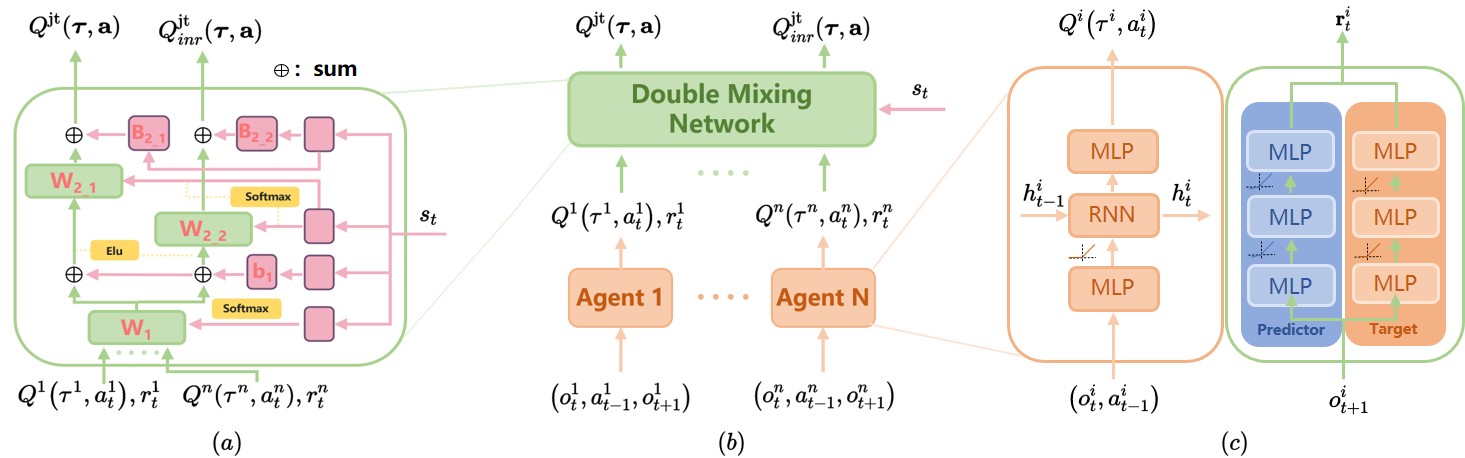}
\caption{$(a)$ Double mixing network structure; $(b)$ SVDE architecture; $(c)$ Agent network, which contains Action network structure (left) and IRD network structure (right), in which predictor network is used to predict next observation, and target network maintains parameters unchanged.}
\label{fig:7}
\vskip -0.15cm
\end{figure*}

The IRD network is shown in Fig.~\ref{fig:7}.
It consists of a target network whose parameters are randomly initialized and fixed in the same way as in RND \cite{rnd}, and a predictor network trained with data collected by the agent.
The fixed parameters of the target network are intended to reduce the randomness of the target fitting function, so that prediction errors are all caused by the novelty of the observation.
Thus, the prediction error is large if similar observations are rarely seen, resulting in a higher intrinsic reward that encourages the agent to explore.

The target and predictor networks accept the same input, which is the local observations of the agent, and output feature vectors of the same dimension: $\hat{f}: \mathcal{O} \rightarrow \mathbb{R}^k, g: \mathcal{O} \rightarrow \mathbb{R}^k$.
We employ weight sharing across the network of agents. 
The agent ID is included in the local observation to allow agents to have different intrinsic rewards.
The intrinsic reward and mean-square error (MSE) loss function are respectively defined as  
\begin{equation}
r_{t}^i=\|\hat{f}(\mathbf{o}_{t+1}^i)-g(\mathbf{o}_{t+1}^i)\|
\label{equ:6}
\end{equation}
\begin{equation}
\mathcal{L}_{inc}(\psi)=\sum_{j=1}^{\mathcal{B}}(\hat{f}(\boldsymbol{\tau}_j; \psi)-g(\boldsymbol{\tau}_j; \psi^-))^2 ,
\label{equ:7}
\end{equation}
where $\mathcal{B}$ is the batch size sampled from the replay buffer, $m$ is a random time step as in DRQN \cite{drqn}, and $l$ is the maximum length of an episode.

We normalize the observations $o_{t+1}^i$ and intrinsic reward $r_{t+1}^i$ to keep the intrinsic reward at a consistent magnitude.
A standard Gaussian distribution is set for observations $obs: \mathbf{N}(0,1)$ and intrinsic rewards $rw: \mathbf{N}(0,1)$, which regularly update the mean and standard deviation based on the network's inputs and outputs: $$\mathbf{o}_{t+1}^i= \frac{o_{t+1}^i-obs.mean}{obs.var}, \mathbf{r}_{t}^i=\frac{r_{t}^i-rw.mean}{rw.var}$$

\label{ird}

\subsubsection{Explorative Experience Replay}
The general off-policy algorithm is uniformly sampled from the replay buffer.
Due to the different quality of the samples, uniform sampling may lead to unnecessary model updates, which reduces   learning efficiency \cite{unnecessary_sampling}.
We propose explorative experience replay to further enhance the utilization of samples at the episode level.
A selection module in the replay buffer, as shown in Fig.~\ref{fig:4}, selects samples according to the \emph{priority} and \emph{importance factor}, and selects the batch data with the highest score for training.
Priority measures the positive effect of the sample on policy convergence.
The importance factor measures the novelty and utilization frequency of the sample.

\textbf{Priority.} 
Due to different temporal difference errors (TD-errors), samples have different effects on backpropagation when updating the network, where TD-errors measure the quality of the update at the current timestep as in QMIX \cite{qmix}.
The algorithm can get more useful information if the TD-error is large.
PER \cite{per} determines the learning potential of the sample according to the TD-errors during the training process.
Following the practice of PER on SARL, we migrate it to MARL.
The replay buffer stores the data in the form of \emph{SumTree} to further improve sampling efficiency and accelerate convergence, as shown in Appendix \ref{sumtree}.

Priority is defined as
\begin{equation}
Priority=\frac{\sum_{t=1}^{l-m}\left[Q^{\mathrm{jt}}(\boldsymbol{\tau}, \mathbf{a}, s ; \theta)-Q_{tar}^{\mathrm{jt}}(\boldsymbol{\tau}, \mathbf{a^{\prime}}, s ; \theta^{-})\right]}{l-m} ,
\label{equ:8}
\end{equation}
where $Q^{\mathrm{jt}}$ and $Q_{tar}^{\mathrm{jt}}$ are the joint Q-function and target Q-function, respectively.
They represent the Q-value and the target Q-value, respectively, of the joint state as in QMIX \cite{qmix}, which we describe later.
The selection module calculates the priority for each sample when it is stored in the replay buffer, and updates its priority after each sampling.
The greater the priority, the more likely the sample is to have a positive effect on policy convergence.

\textbf{Importance factor.} 
Since PER stores a transition, it does not fully consider the inherent characteristics of the sample distribution in the replay buffer.
In contrast, we store an episode with cumulative external rewards $\mathcal{R}^e=\sum_{t=0}^{l} r_{t}^e$, maximum length $l$, and number of visits $\mathcal{N}$.
$\mathcal{R}^e$ and $l$ can evaluate the quality of the episode.
A large $\mathcal{R}^e$ and a small $l$ indicate that the episode is of high quality, i.e., the agent has obtained a high reward with very few timesteps.
$\mathcal{N}$ can measure how often the episode is used.
Additionally, we consider the difference between the sample generation round $T_{gen}$ and the current training round $T_{now}$, $\delta=T_{now}-T_{gen}$.
The larger the $\delta$, the more likely the sample is out of date, and newer samples should be appropriately selected.
As $\mathcal{N}$ and $\delta$ increase, the importance factor gradually decreases, which means that the sample is used more often or is outdated and a newer sample should be selected:

\begin{equation}
Importance\,factor=\max (\frac{\mathcal{R}^e}{l} + \mathcal{C} \cdot \delta \cdot \sqrt{\ln \mathcal{N}}, 0 ) ,
\label{equ:9}
\end{equation}
where the first item in the first parameter of the max function ensures the initial value of an episode, and the second item in the first parameter of the max function ensures exploration.
$\mathcal{C}$ is a negative constant, and the importance factor is greater than or equal to $0$, as shown in \eqref{equ:9}.

\label{eeif}

\subsection{Network Architecture}
Fig.~\ref{fig:7} shows the SVDE architecture.
To decompose the optimal joint Q-function and generalize the representation to a larger class of monotonic functions, the joint Q-function must satisfy \eqref{equ:3}.
The architecture of SVDE follows this principle.

For each agent $i$, the agent network has two parts: the action network and the IRD network.
The action network is a deep recurrent Q-network that receives the local action-observation history $\tau^i$, i.e., current local observations $o_t^i$, last local actions $a_{t-1}^i$, and last hidden states $h_{t-1}^i $, as input.
The IRD network is a feedforward neural network that receives next local observations $o_{t+1}^i$ as input.
The agent network outputs the local Q value $Q^i\left(\tau^i, a^i\right)$ and intrinsic reward $\mathbf{r}_{t}^i$ at each time step.
We employ weight sharing among agents, so we train only one agent network for all agents, including the action and IRD networks.
The ID of the agents is input to the network as part of the local observation, so that each agent has potentially different actions and obtains different intrinsic rewards.

The double mixing network is a feedforward neural network that takes the output $Q^i\left(\tau^i, a_t^i\right)$ of the agent network as input, and mixes them monotonically to output joint Q-value $Q^{\mathrm{jt}}$ and intrinsic joint Q-value $Q_{inc}^{\mathrm{jt}}$, where $Q_{inc}^{\mathrm{jt}}$ is trained entirely using intrinsic rewards.
To satisfy \eqref{equ:3}, the weights (excluding bias) of the feedforward network must be nonnegative.
This allows the double mixing network to arbitrarily approximate any monotonic function.

The weights of the double mixing network are generated by the hypernetworks, which receive the global state $s$ as input and generate the weights of one layer of the double mixing network.
Each hypernetwork consists of a linear layer whose output is fed into a \emph{softmax} layer to ensure that the weights are nonnegative.
However, the biases do not need to be nonnegative, so there is no need for a \emph{softmax} layer behind the corresponding hypernetwork.
The first layer $W_1, b_1$ of the double mixing network is a shared layer, and the second layer has two branches with weights generated by different hypernetworks.
The branch $W_{2\_1}, b_{2\_1}$ is used to calculate $Q^{\mathrm{jt}}$, and the branch $W_{2\_2}, b_{2\_2}$ is used to calculate $Q_{inc}^{\mathrm{jt}}$.

To prevent the agent from forgetting the original goal because of the high intrinsic reward, we consider the double mixing network output joint Q-value $Q^{\mathrm{jt}}$ and intrinsic joint Q-value $Q_{inc}^{\mathrm{jt}}$, whose TD-errors are calculated as
\begin{equation}
\delta_{ext}=y^{\mathrm{jt}}-Q^{\mathrm{jt}}(\boldsymbol{\tau}, \mathbf{a}, s ; \theta)
\label{equ:10}
\end{equation}
\begin{equation}
\delta_{inc}=y_{inc}^{\mathrm{jt}}-Q_{inc}^{\mathrm{jt}}(\boldsymbol{\tau}, \mathbf{a}, s ; \theta_{inc})
\label{equ:11} ,
\end{equation}
where $y^{\mathrm{jt}}=\mathbf{r}+\gamma_1\max_{\mathbf{a}^\prime}Q^{\mathrm{jt}}(\boldsymbol{\tau}^\prime, \mathbf{a}^\prime, s^\prime ; \theta^-)$, $y_{inc}^{\mathrm{jt}}=\mathbf{r}^i+\gamma_2\max_{\mathbf{a}^\prime}Q_{inc}^{\mathrm{jt}}(\boldsymbol{\tau}^\prime, \mathbf{a}^\prime, s^\prime ; \theta^-)$, and $\theta^-$ denotes the parameters of the target network.

We define \emph{ISweight} (abbreviated as \emph{IS}), which reflects the importance of the current sample in terms of priority and importance factor. 
It is calculated as 
\begin{equation}
IS=\alpha \cdot Priority + (1-\alpha) \cdot Importance \, factor ,
\label{equ:12}
\end{equation}
where $\alpha \in [0,1]$ is a hyperparameter.
If \emph{IS} is large, it indicates that the sample's quality is high after weighing it from two perspectives. 

The loss of the double mixing network is defined as    
\begin{equation}
\mathcal{L}_{mix}(\theta, \theta_{inc})= \\
\sum_{j=1}^{\mathcal{B}}\left[IS^j \left((1-\beta) \cdot \delta_{ext}^j 
+ \beta \cdot \delta_{inc}^j\right)^2\right] ,
\label{equ:13}
\end{equation}
where $\beta \in [0,1]$ is a hyperparameter used to trade off exploration.
$\beta$ is set to a higher value at the start of a game, allowing agents to explore environments as much as possible. 
It gradually decreases to 0 as the game progresses in order to prevent the agent from blindly obtaining intrinsic rewards and forgetting the ultimate goal, which means less exploration in the later stages of a game.

SVDE is trained end-to-end to minimize:

\begin{equation}
\mathcal{L}(\theta, \theta_{inc}, \psi)=\mathcal{L}_{mix}(\theta, \theta_{inc})+\mathcal{L}_{inc}(\psi) ,
\label{equ:14}
\end{equation}
where the first item is the loss function of the double mixing network, and the second item is the MSE loss function of the IRD network as in \eqref{equ:7}.

\label{method}

\section{Experiments}
We verify the effectiveness of SVDE for \emph{data-efficiency} and \emph{exploration and exploitation} in MARL on the StarCraft Multi-Agent Challenge (SMAC) platform; analyze the scalable training mechanism, intrinsic reward design, and explorative experience replay; and describe ablation studies on the effects of the main components in our method.

\begin{figure}[!h]
    \begin{minipage}[t]{1\linewidth}
    \centering
    \subfloat[]{
    \includegraphics[width=8cm]{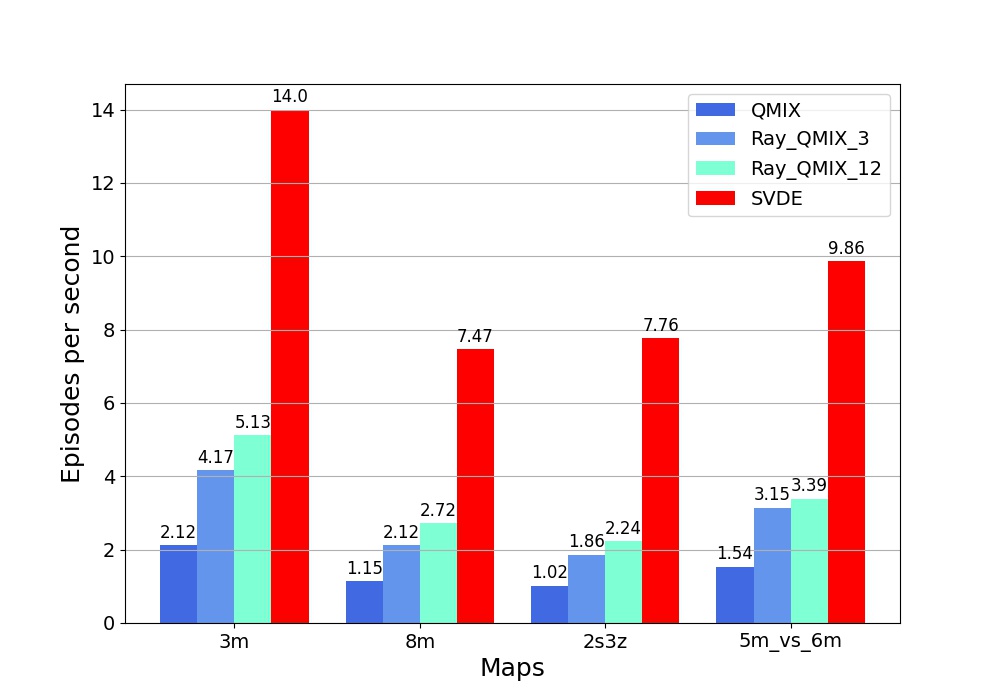}
    \label{subfig:3}
    }
    \end{minipage}
    \vskip -0.1cm
    \begin{minipage}[t]{1\linewidth}
    \centering
    \subfloat[]{
    \includegraphics[width=8cm]{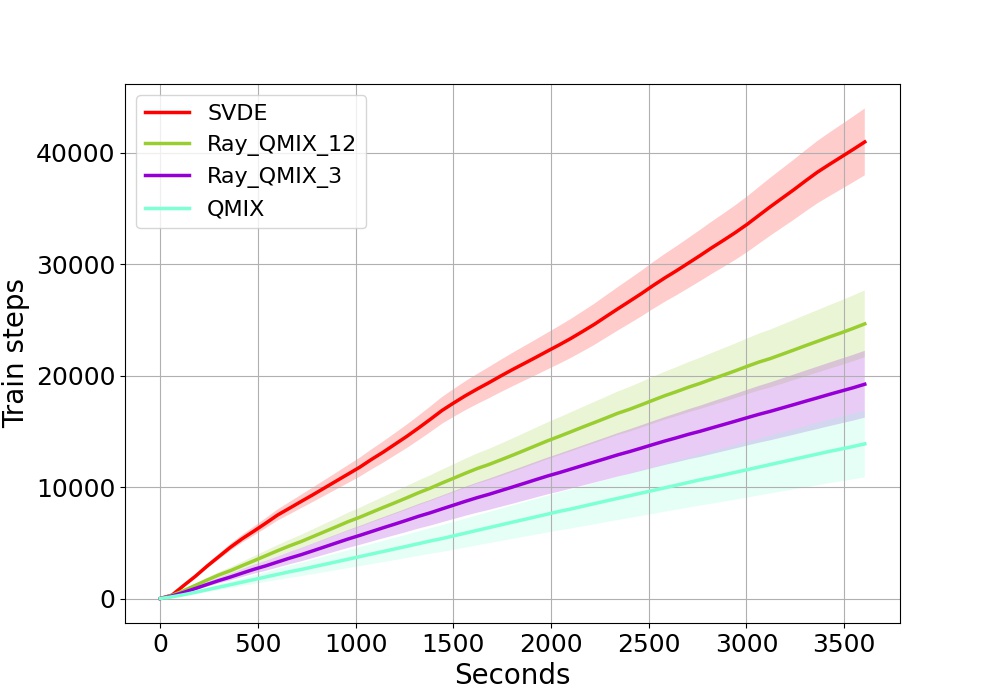}
    \label{subfig:4}
    }
    \end{minipage}
\caption{$(a)$ Sample collection and training speed; $(b)$ Ray$\_$QMIX$\_$3 represents QMIX using Ray, and sets the workers to three. Ray$\_$QMIX$\_$12 sets the workers to 12. SVDE sets three workers, each of which sets four actors.}
\label{fig:8}
\vskip -0.5cm 
\end{figure}

\begin{figure*}[t]
    \centering
    \vskip -0.35cm
    \begin{minipage}[t]{1\linewidth}
        \centering
        \subfloat[]{
        \includegraphics[width=5.85cm]{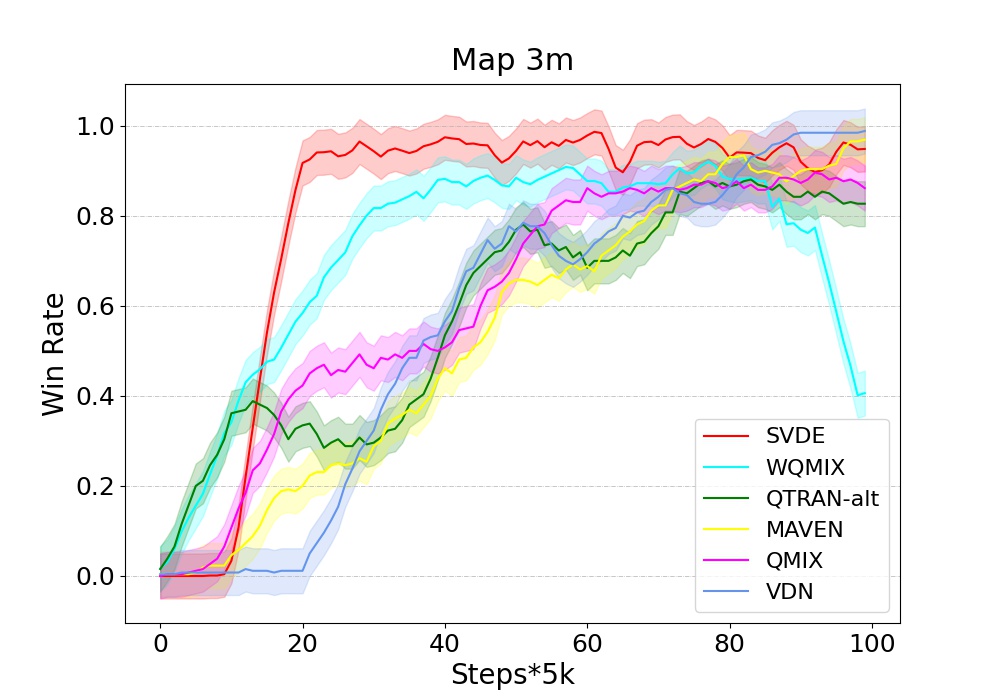}
        \label{subfig:5}
        }\noindent
        \subfloat[]{
        \includegraphics[width=5.85cm]{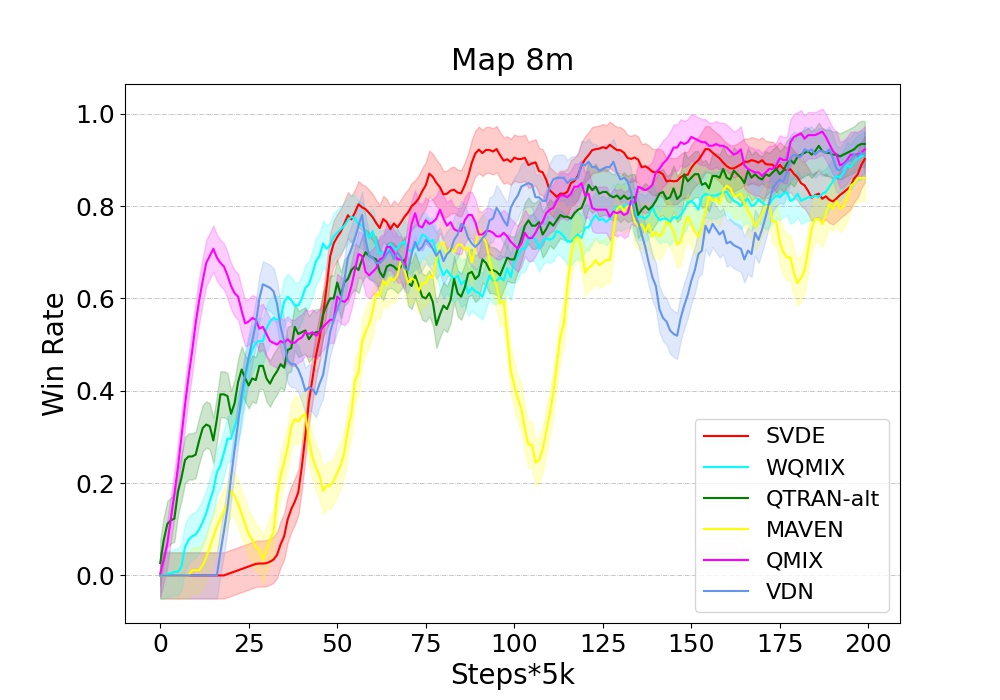}
        \label{subfig:6}
        }\noindent
        \subfloat[]{
        \includegraphics[width=5.85cm]{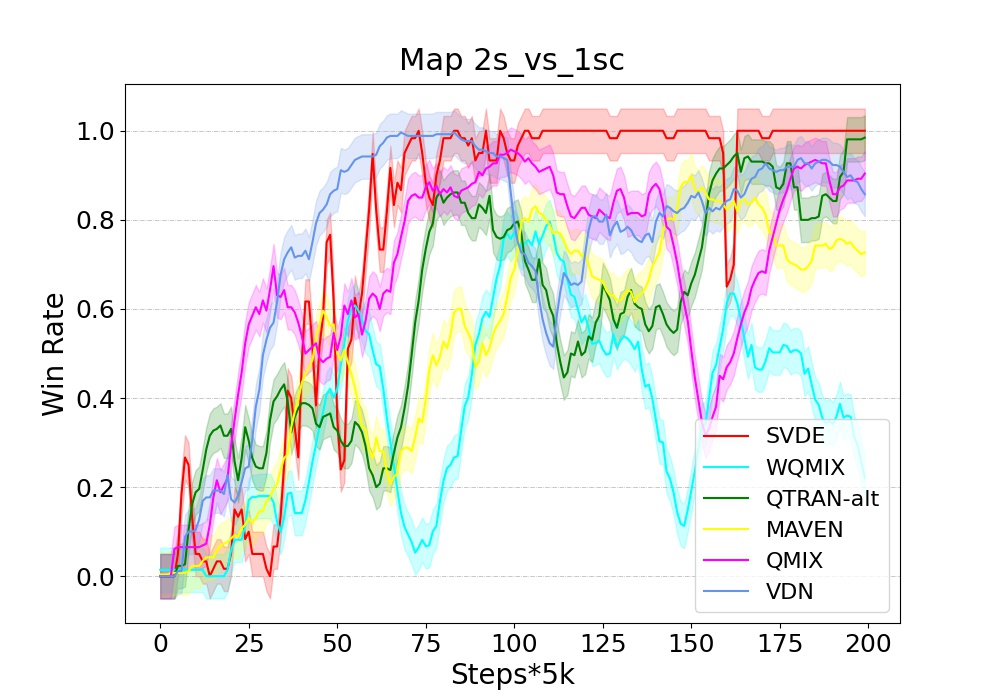}
        \label{subfig:7}
        }
    \end{minipage}%
    \vskip -0.15cm
    \begin{minipage}[t]{1\linewidth}
        \centering
        \subfloat[]{
        \includegraphics[width=5.85cm]{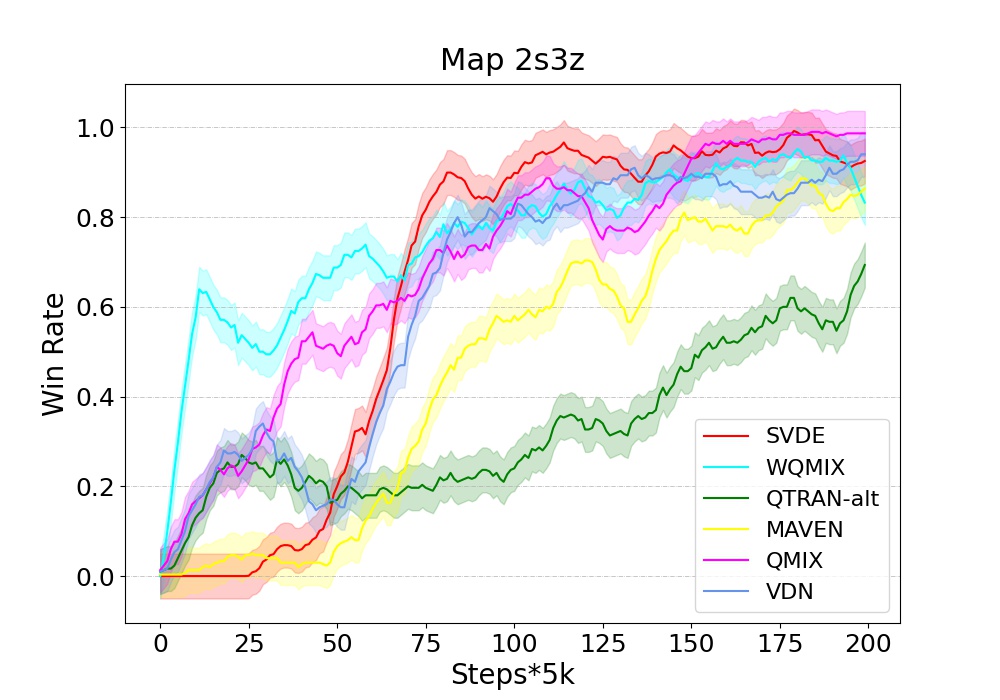}
        \label{subfig:8}
        }\noindent
        \subfloat[]{
        \includegraphics[width=5.85cm]{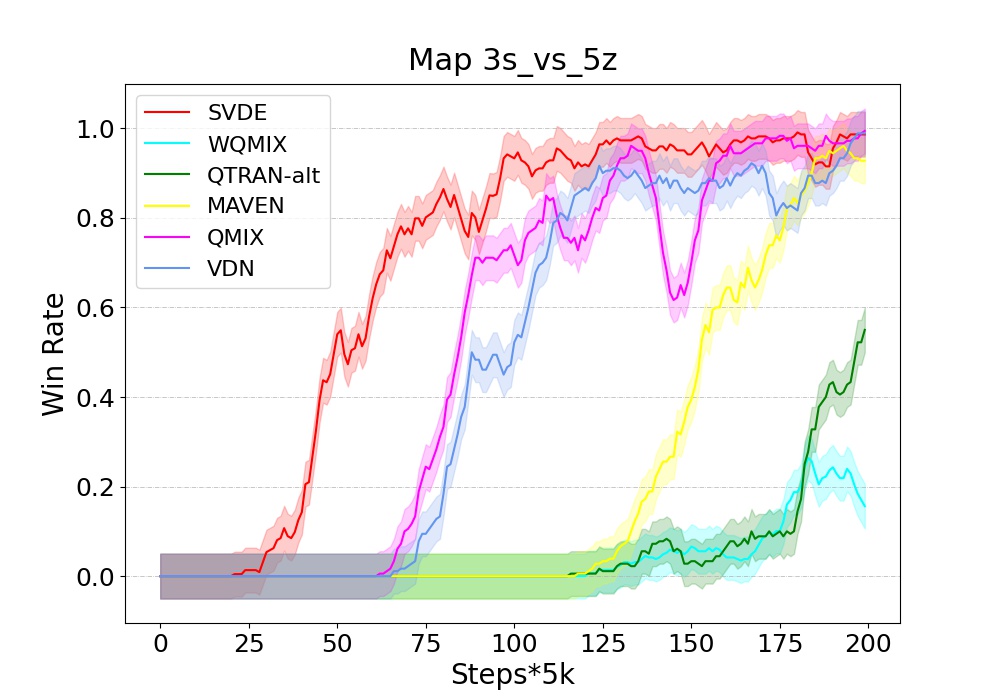}
        \label{subfig:9}
        }\noindent
        \subfloat[]{
        \includegraphics[width=5.85cm]{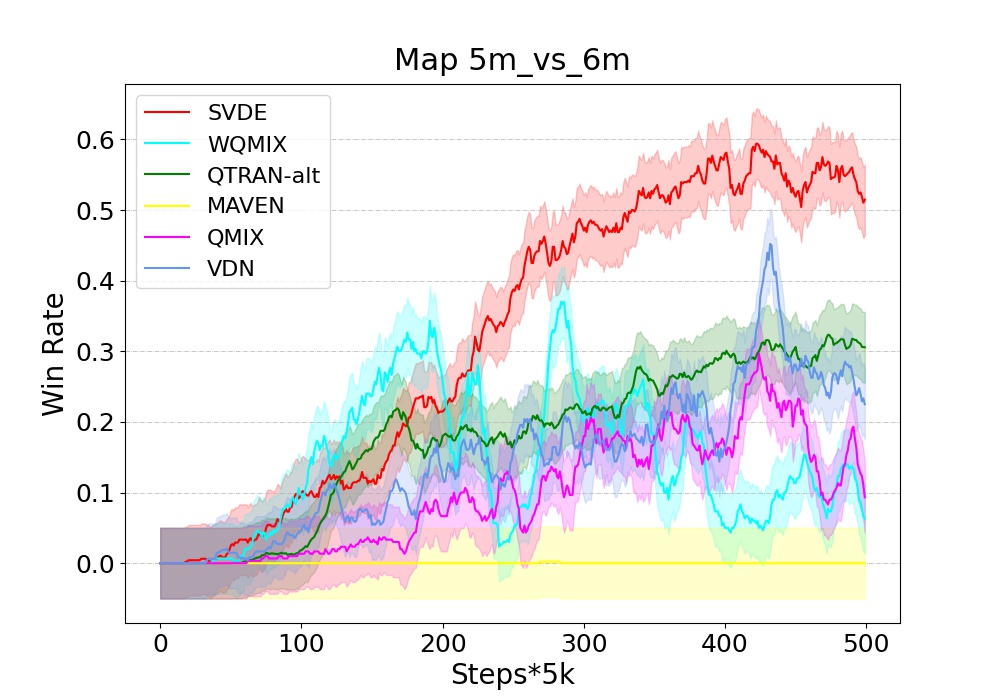}
        \label{subfig:10}
        }
    \end{minipage}%
\caption{\textbf{Overall results:} Win rate on six maps.}
\label{fig:9}
\end{figure*}

\subsection{Data-efficiency}
Data-efficiency is a bottleneck restricting RL algorithms \cite{ray, mava}.
Our proposed SVDE method has a scalable training mechanism that can sample massively in parallel.
Moreover, the training and rollout processes are completely decoupled and independent, which accelerates training and convergence.
We compare SVDE with the popular baseline algorithm QMIX \cite{qmix} and distributed training framework Ray \cite{ray}, with results as shown in Fig.~\ref{fig:8}.
Appendix \ref{smac} shows details of the experimental maps.

Under the same three-worker setting, Ray has an acceleration effect comparable to QMIX.
However, the improvement effect is not ideal in the 12-worker setting.
This is because the experimental machine only has two physical CPUs (Appendix \ref{training configurations}), and the structure of Ray dictates that it needs high computing power and complex remote procedure call (RPC) to enable its distributed capabilities \cite{ray}.
Although SVDE also has three workers, it has 12 parallel actors due to the design pattern, and it has almost linear incremental sampling efficiency.
Due to the different functional designs of the actor and worker, the actor does not need to perform serving but just interact with the environment.
It reduces the computational complexity of actors and speeds up sampling.
Under the same parallel settings, Ray needs 12 workers for serving, and SVDE only needs three.
SVDE has better distributed features than Ray, and it requires fewer computing resources.

\subsection{Decentralized Micromanagement Tasks}
We evaluated SVDE on the StarCraft II decentralized micromanagement tasks, where each learning agent controls an individual army unit.
We used the SMAC environment \cite{smac} as our test platform; this is a common benchmark for evaluating state-of-the-art MARL methods such as VDN \cite{vdn}, QMIX \cite{qmix}, QTRAN \cite{qtran}, WQMIX \cite{wqmix}, and MAVEN \cite{maven}.
On the SMAC platform, two armies battle in each map scenario.
We trained multiple agents to control allied units to defeat enemies, whose units were controlled by a built-in handcrafted AI.
An episode ended when all units of either army had died or the episode had reached a predefined time limit.
The win condition was that all enemy units were eliminated.

We used the following recording module to assess how well each method performed: 
For each run of each method, we pause the network training of the learner every 5000 (5k) steps and run 32 independent episodes with each agent following the current policy to choose an action greedily. 
The win rate is the proportion of these episodes in which the method defeats all enemy units in the allotted amount of time.

Considering the characteristics and difficulty of maps (Appendix \ref{smac}), we selected six maps to make the experiment cover all scenarios as much as possible: $3m$, $8m$, $2s_{-} vs_{-} 1sc$, $2s3z$, $3s_{-} vs_{-} 5z$, and $5m_{-} vs_{-} 6m$.
Appendix \ref{smac} shows the map details, and hyperparameter settings used in the experiments are shown in Appendix \ref{hyper}.

\begin{table*}[t]
    \centering
    \caption{Median performance of test win rate on experimental maps.}
    \setlength{\tabcolsep}{7mm}
    \begin{tabular}{ccccccc}
        \hline
         Method & $3m$ & $8m$ & $2s_{-} vs_{-} 1sc$ & $2s3z$ & $3s_{-} vs_{-} 5z$ & $5 m_{-} v s_{-} 6 m$ \\
        \hline
        VDN & \textbf{0.95} & 0.85 & 0.95 & 0.8 & 0.85 & 0.19\\
        QMIX & 0.9 & 0.85 & 0.9 & \textbf{0.9} & 0.9 & 0.05\\
        QTRAN-ALT & 0.9 & 0.85 & 0.9 & 0.45 & 0.05 & 0.25\\
        WQMIX & 0.91 & 0.8 & 0.66 & 0.81 & 0 & 0.1\\
        MAVEN & 0.9 & 0.75 & 0.7 & 0.75 & 0.38 & 0\\
        \textbf{SVDE (Ours)} & 0.93 & \textbf{0.93} & \textbf{1.0} & \textbf{0.9} & \textbf{0.91} & \textbf{0.47} \\
        \hline
    \end{tabular}
    \vskip -0.25cm
    \label{tab:1}
\end{table*}

\begin{figure*}[t]
    \begin{minipage}[t]{1\linewidth}
    \centering
    \subfloat[]{
    \includegraphics[width=5.85cm]{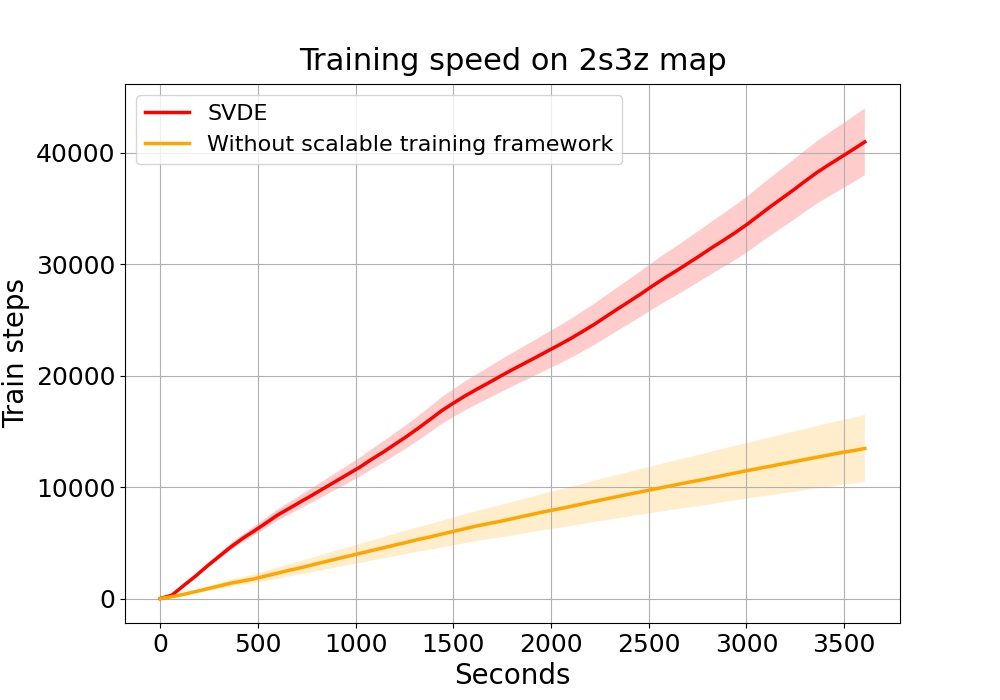}
    \label{subfig:11}
    }\noindent
    \subfloat[]{
    \includegraphics[width=5.85cm]{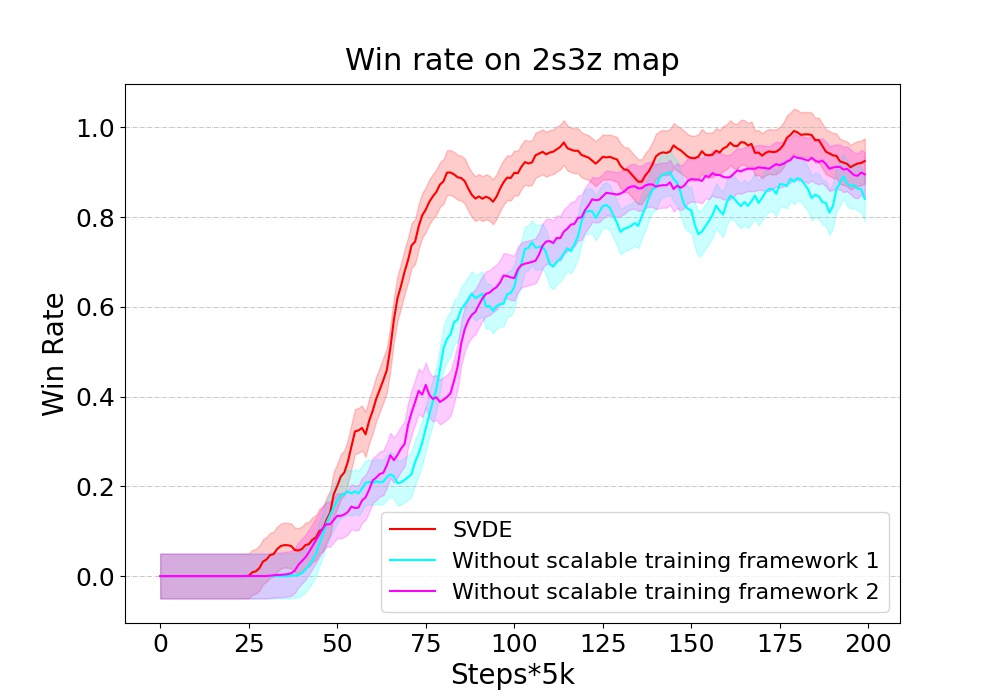}
    \label{subfig:12}
    }\noindent
    \subfloat[]{
    \includegraphics[width=5.85cm]{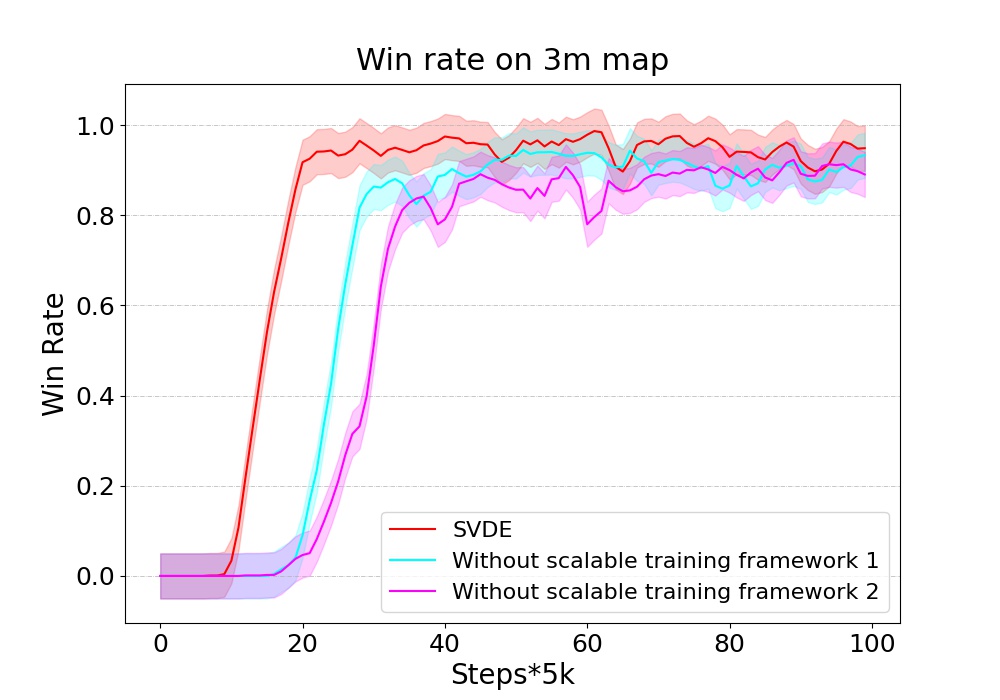}
    \label{subfig:13}
    }
    \end{minipage}
    \vskip -0.15cm
    \begin{minipage}[t]{1\linewidth}
    \centering
    \subfloat[]{
    \includegraphics[width=5.85cm]{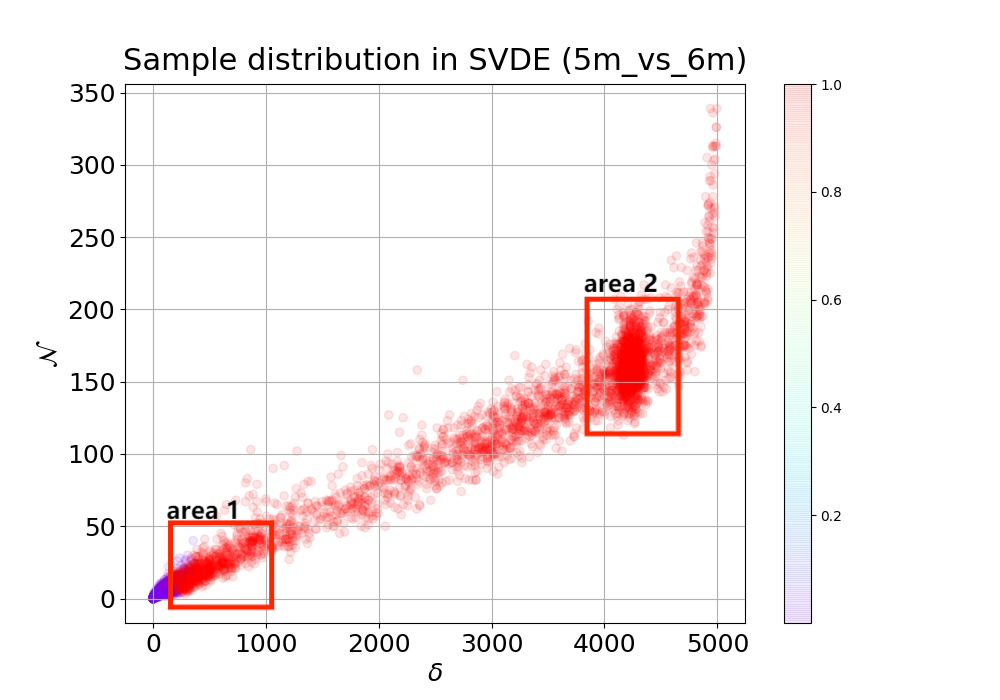}
    \label{subfig:14}
    }\noindent
    \subfloat[]{
    \includegraphics[width=5.85cm]{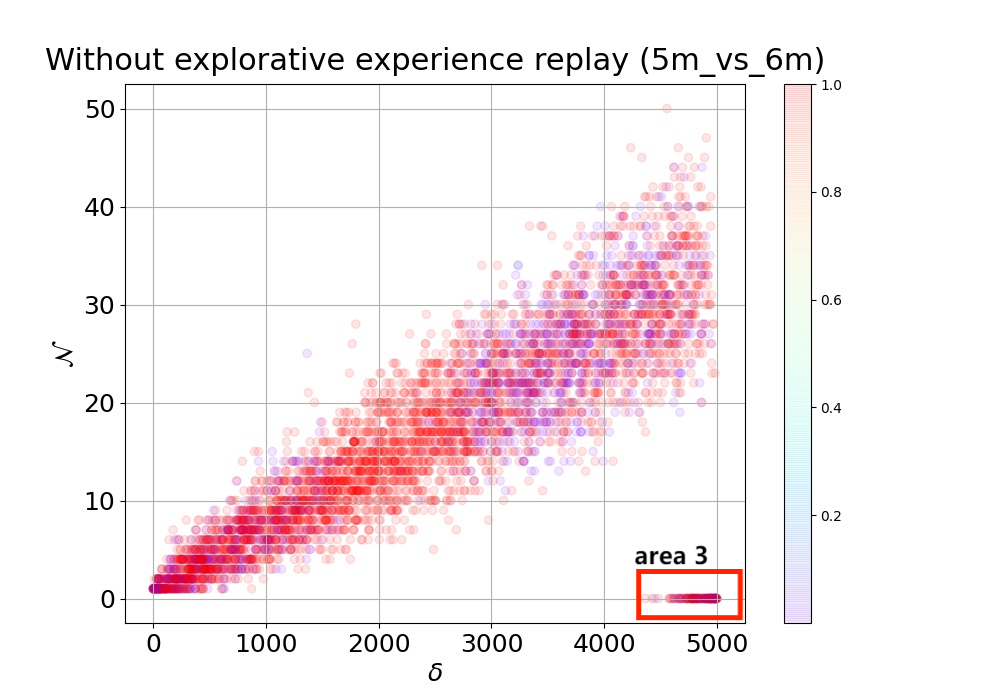}
    \label{subfig:15}
    }\noindent
    \subfloat[]{
    \includegraphics[width=5.85cm]{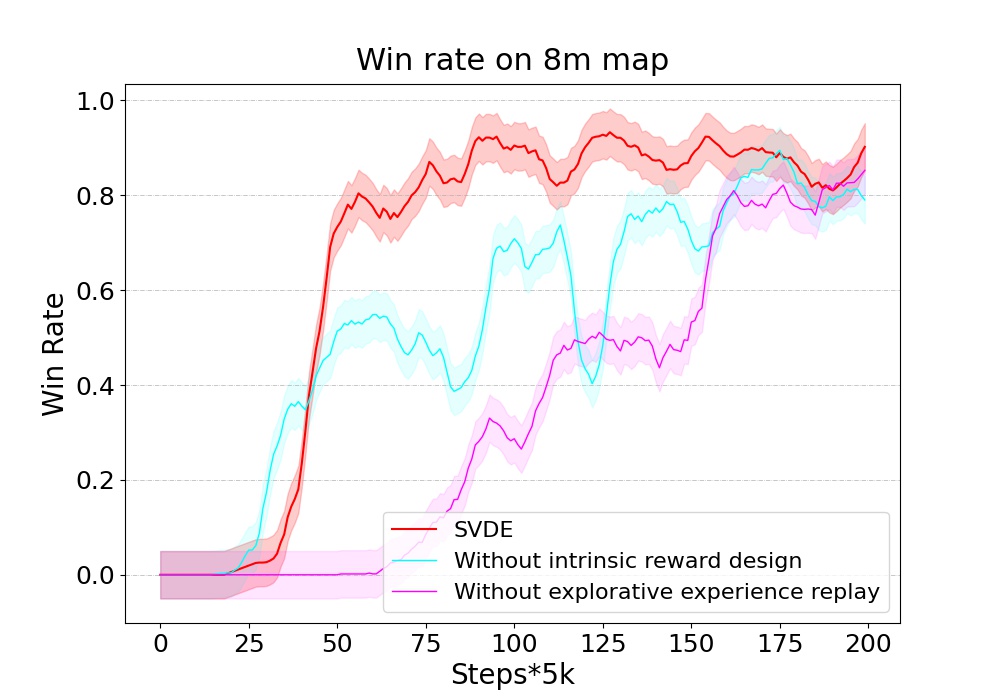}
    \label{subfig:16}
    }
    \end{minipage}
\caption{\textbf{Ablation:} $(a)$, $(b)$, and $(c)$ are \textbf{\emph{ablation 1}}. $(d)$, $(e)$, and $(f)$ are \textbf{\emph{ablation 2}}. $(d)$ and $(e)$ represent the sample distribution in the replay buffer, which has 5,000 samples in total, and the color represents the importance factor (equation \eqref{equ:9}) of the sample.}
\label{fig:10}
\vskip -0.15cm
\end{figure*}

Using the MARL algorithms as a baseline, the win rates on the six maps are shown in Fig.~\ref{fig:9}.
On simple maps such as $3m$ and $8m$, almost all algorithms performed well.
However, the win rate curve of SVDE rose steadily over time, and eventually stabilized at a higher win rate than the other algorithms.
On difficult maps (e.g., $2 s_{-} v s_{-} 1 sc$), it had a longer exploration cycle due to its map settings.
Almost all the other methods had large performance fluctuations, while SVDE had a small fluctuation in early exploration, but gradually stabilized and converged to a win rate of 100\% after enough exploration.
All other methods were prone to fall into local optima without active exploration, and uniform sampling might cause unnecessary updates that have the opposite effect on policy improvement.
SVDE converged more easily to the global optimum due to sufficient exploration.

WQMIX was the fastest method to reach a win rate of 70\%, and QTRAN did not learn an effective strategy on $2s3z$ due to its high complexity.
Because of its long exploration period on $3 s_{-} v s_{-} 5 z$, SVDE became the fastest converging method, and it achieved the best performance.
On the most difficult map, $5 m_{-} vs_{-} 6m$, all other methods failed to learn effective strategies, and their maximum win rate was around 40\%, while SVDE achieved a win rate of around 60\%.

In summary, SVDE may not converge fastest because it needs to explore the environment.
But the proposed scalable training mechanism compensates for the time spent on exploration.
SVDE performs effective exploration and exploitation, and is best on almost all experimental maps.

\subsection{Ablation}
To demonstrate the effectiveness of SVDE, we performed \emph{ablation 1} to prove the necessity of the scalable training mechanism for training acceleration and convergence, and \emph{ablation 2} to demonstrate the importance of the intrinsic reward design and explorative experience replay.

\subsubsection{\textbf{Ablation 1}}
Fig.~\subref*{subfig:11} shows that when the scalable training mechanism is not used, only about 12,000 training times can be performed in one hour, and 40,000 can be performed after using it, an improvement of almost 3.5 times.
Fig.~\subref*{subfig:12} and Fig.~\subref*{subfig:13} show the acceleration of convergence of the method due to  distributed training, which shows the effectiveness of the scalable training mechanism.
Fig.~\subref*{subfig:12} shows that when the abscissa reaches 75, SVDE has converged and has a win rate of around 90\%, compared to a win rate of around 50\% for the control group.
Additionally, in Fig.~\subref*{subfig:13}, SVDE already has a win rate of around 95\% when the abscissa is 20, while the control group only has a win rate around 0.

\subsubsection{\textbf{Ablation 2}}
When uniform sampling is used, samples with different importance factors are evenly distributed in the replay buffer, and they are randomly sampled, as shown in Fig.~\subref*{subfig:15}.
They may not be sampled many times, and some samples will not be sampled all the time, even if the importance factor is large (area 3 in Fig.~\subref*{subfig:15}).

In contrast, after adopting explorative experience replay, the samples are mainly distributed in two areas, as shown in Fig.~\subref*{subfig:14}.
Some new samples with large importance factors have not been sampled multiple times and gather near the origin (area 1 in Fig.~\subref*{subfig:14}), while other samples with large importance factors are clustered in the area where number of visits $\mathcal{N}$ is large (area 2 in Fig.~\subref*{subfig:14}), showing that the algorithm has purposefully sampled them multiple times for training.
Furthermore, explorative experience replay makes better sample utilization nearly 300 times (the ordinate of Fig.~\subref*{subfig:14}), which is a nearly 6-fold improvement compared to 50 times without it (the ordinate of Fig.~\subref*{subfig:15}).
It ensures that every sample is sampled, and there is no sample with a visit count $\mathcal{N}$ of 0 (bottom right corner of Fig.~\subref*{subfig:15}).

Fig.~\subref*{subfig:16} shows the win rate on the $8m$ map.
Although no exploration will speed up the learning of the Q-function, it will cause the process to become unstable and eventually converge to a local optimal joint Q-function.
Moreover, inability to effectively use samples that have large importance factors decelerates the learning process, making convergence slower.

\section{Conclusion}
We proposed a value-decomposition method based on scalable exploration for data-efficiency and exploration-exploitation problems in MARL.
We first proposed a scalable training mechanism for the lack of sample generation, and divided the learning process into the three processes of rollout, serving, and training, which are suitable for cooperative MARL.
We generated a large number of samples, and used intrinsic reward design and explorative experience replay to generate better-quality samples, which we used to accelerate the convergence of training.
Empirically, our method performed best compared to other popular algorithms on almost all maps in challenging StarCraft II micromanagement games.
We demonstrated the effectiveness of our method through ablation experiments.

Since all experiments were completed on a single-node machine with multiple GPUs of the same type, the current scalable training mechanism does not consider the waiting and synchronization issues caused by the computing powers of different GPUs in a multi-node cluster.
Furthermore, we have not yet considered the communication mechanism between multiple agents, so each agent independently chooses the local optimal action instead of trying to communicate to choose the globally optimal action.

In future work, we aim to combine communication methods to realize the informational communication of agents and expand the scalable mechanism to distributed clusters, so as to implement our method in real-world applications.


\bibliographystyle{ieeetr}
\bibliography{work_1}

\appendix
\subsection{SumTree}
\begin{figure}[htbp]
    \centering
    \includegraphics[width=8cm]{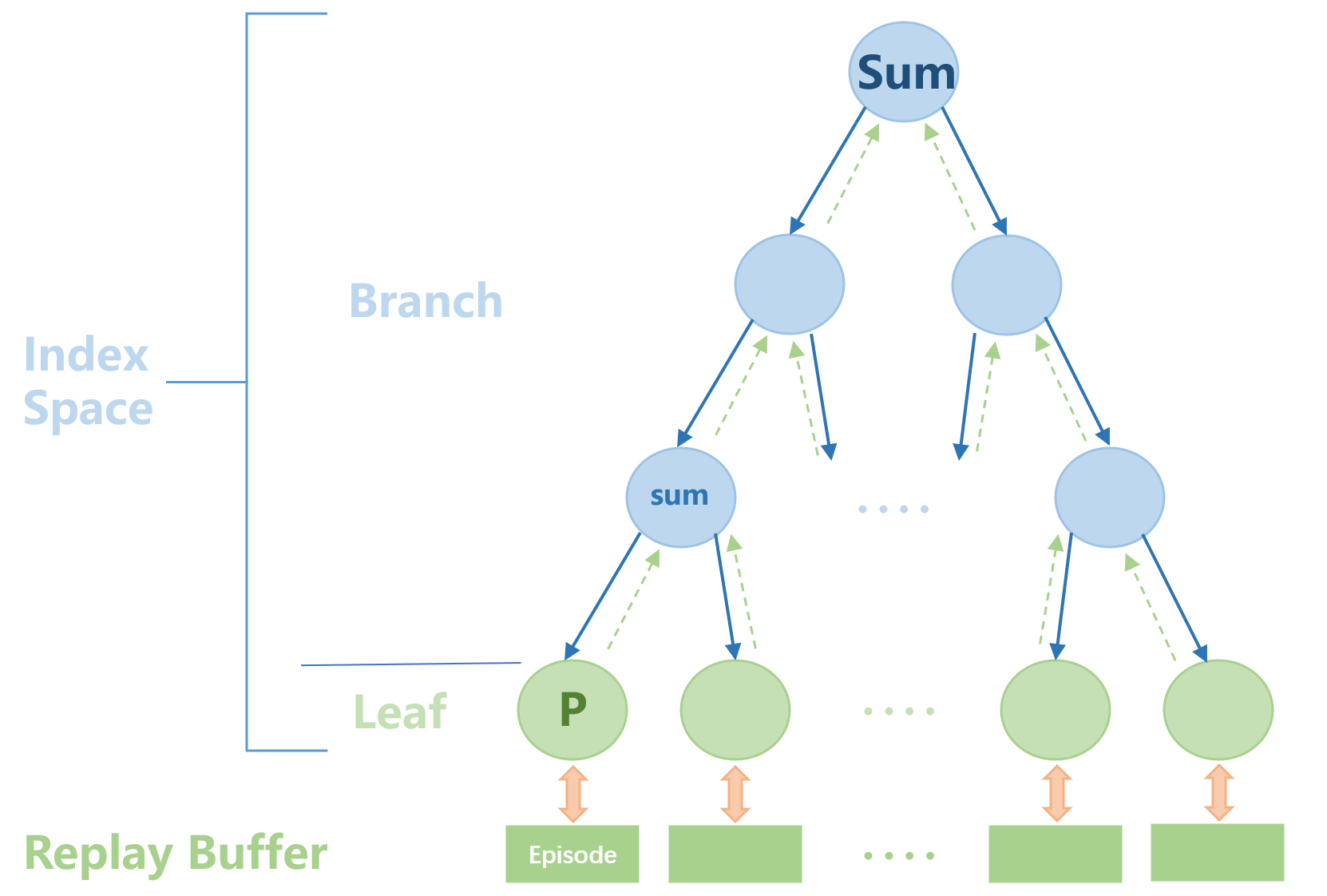}
    \caption{SumTree architecture.}
    \label{fig:x}
\end{figure}

The SumTree is a tree structure that consists of two spaces. 
Each leaf in the index space stores the priority $P$ of a sample, while each branch node has only two forks, and the value of the node is the sum of the two forks. 
Therefore, the top node of the SumTree is the sum of the priorities of all nodes. 
The second space is the replay buffer, which is used to store sample data. In this space, each node stores a sample corresponding to the priority in the index space.
The structure of SumTree is shown in Fig.~\ref{fig:x}.

The priority $P$ is calculated according to \eqref{equ:12}.
The SumTree data structure is divided into two processes: sampling and updating. 
During the sampling process, the capacity of the replay buffer is divided by the batch size to obtain intervals, and uniform random sampling is performed in each interval using \eqref{equ:16}, where $\mathcal{M}$ is the capacity of the replay buffer and $l$ is the batch size. 
\begin{equation}
Interval=\frac{\mathcal{M}}{l}
\label{equ:15}
\end{equation}
\begin{equation}
episode_i \sim uniform(Interval_i)\quad for\,i \in 1 \rightarrow l 
\label{equ:16}
\end{equation}

In this way, the episode with large priority is easier to be sampled, as shown by the solid line in Fig.~\ref{fig:x}.
The updating process is performed after each training process.
First update the priority of the corresponding leaf node according to \eqref{equ:12}, and then backpropagate to the ancestor branch nodes for a layer-by-layer update, as shown by the dotted line in Fig.~\ref{fig:x}.

\label{sumtree}

\subsection{SMAC}
The difficulty on the SMAC platform can be customized, and we use the ”very difficult” mode, which involves setting the game difficulty to level 7 and adjusting parameters such as shooting range and observation range.
Depending on whether the number of units is the same or not, battle maps can be categorized into symmetric or asymmetric battle maps, as well as homogeneous or heterogeneous battle maps based on the unit types on both sides.

\subsubsection{Map Settings}
For the SMAC experiment, we selected six maps, which are $3m$, $8m$, $2s_{-} vs_{-} 1sc$, $2s3z$, $3s_{-} vs_{-} 5z$, and $5m_{-} vs_{-} 6m$.
Table~\ref{tab:2} provides details on each map.
\begin{table}[h]
    \centering
    \caption{Map Descriptions. We use $IS$ for isomorphic symmetry, $HES$ for heterogeneous symmetry, $HOA$ for homogeneous asymmetric, $HEA$ for heterogeneous asymmetric, $M$ for Marines, $S$ for Stalkers, $Z$ for Zealots, $SC$ for Spine Crawler.}
    \begin{tabular}{c|c|c|c}
        \hline
         Map Name &  Own Units & Enemy Units & Map Type \\
        \hline
        $3m$ & $3 M$ & $3 M$ & $IS$\\
        $8m$ & $8 M$ & $8 M$ & $IS$\\
        $2s_{-} vs_{-} 1sc$ & $2 S$ & $1 SC$ & $HEA$\\
        $2s3z$ & $2 S$ and $3 Z$ & $2 S$ and $3 Z$ & $HES$\\
        $3s_{-} vs_{-} 5z$ & $3 S$ and $5 Z$ & $3 S$ and $5 Z$ & $HES$\\
        $5 m_{-} v s_{-} 6 m$ & $5 M$ & $6 M$ & $HOA$ \\
        \hline
    \end{tabular}
    \label{tab:2}
\end{table}

Similar to the work of \cite{qmix,coma}, the action space of agents consists of the following set of discrete actions: \text{move[direction]}, \text{attack[enemy id]}, \text{stop}, and \text{no operation}.
Agents can only move in the following four directions: \text{East}, \text{South}, \text{West}, and \text{North}.
The units are only allowed to take the attack[enemy id] action if the enemy is within shooting range.
As healer units, medivacs use heal[agent id] actions instead of attack[enemy id].
Moreover, each unit has a pre-defined \text{sight range} according to the map, which is used to limit the range of any information they receive, whether it is from allies or enemies.
A unit can only observe other units when they are both alive and in sight range, i.e., the unit cannot distinguish whether other units are dead or out of sight range.

\begin{table*}[t]
    \centering
    \caption{The feature information of each map.}
    \begin{tabular}{ccccccc}
        \hline
         Map &  Action Number & Ally Unit Number & Enemy Unit Number &State Dimension & Observation Dimension & Time-step limit \\
        \hline
        $3m$ & 9 & 3 & 3 & 48 & 30 & 60 \\
        $8m$ & 14 & 8 & 8 & 168 & 80 & 120\\
        $2s_{-} vs_{-} 1sc$ & 7 & 2 & 1 & 27 & 17 & 300\\
        $2s3z$ & 11 & 2 & 3 & 120 & 80 & 120\\
        $3s_{-} vs_{-} 5z$ & 11 & 3 & 5 & 68 & 48 & 250\\
        $5 m_{-} v s_{-} 6 m$ & 12 & 5 & 6 & 98 & 55 & 70\\
        \hline
    \end{tabular}
    \label{tab:3}
\end{table*}
\subsubsection{Observations and States}
The observation vector contains the following properties for allied and enemy units within sight range: distance, relative x, relative y, health, shield, and unit type.
Additionally, it includes the last actions of allied units in sight range, the terrain features, and surrounding units within sight range.
The global state vector $s$ encodes information about all units on the map and is only provided during centralized training.
It specifically includes the coordinates of all units relative to the center of the map and the observed feature of all units that is in observation.
Furthermore, the state stores the power/cooldown of allied units based on unit attributes, which represent the minimum delay between attacks/heals.
Finally, the last actions of all units are appended to the state vector.
See Table~\ref{tab:3} for specific information.

\subsubsection{Rewards}
At each time step, the agent gets a team reward that is equal to the total damage done to enemy units.
Moreover, the agent is rewarded with 10 points after killing each enemy unit and 200 points after killing all enemies for winning the battle.
The rewards are scaled so that the maximum cumulative reward in each map is approximately 20.

\label{smac}

\subsection{Training Configurations}
Experiments are obtained by using GPU \emph{Quadro RTX 8000} and CPU \emph{Intel(R) Xeon(R) Gold 6150}.
Each independent run takes 1–5 hours, depending on the agent numbers and map features of each map.
Each independent run corresponds to a random seed that generalizes randomly before starting.

The action network is designed using DRQN \cite{drqn}, and its recurrent layer consists of a GRU layer with a 64-dimensional hidden state, with a fully-connected layer before and after.
The IRD network consists of two network blocks, predictor and target, and each network block contains three fully-connected layers.
Its hidden layer has a 32-dimensional hidden state, and the output state is fixed as a 5-dimensional vector.
The hypernetworks in the double mixing network employ a 32-dimensional hidden state.

The number of the total training steps is about 1 million ($3m$ 0.5 million, $5 m_{-} v s_{-} 6 m$ 2.5 million) and every 5000 steps we evaluate the model.
The replay buffer stores the latest 5000 episodes, and 32 episodes will be sampled from it before each training.
The target networks are updated after every 200 training times.
In all experiments, the learner of SVDE sets three workers, each of which sets four actors.

\label{training configurations}

\subsection{Hyperparameter Settings}
The hyperparameter settings in the experiment are shown in Table~\ref{tab:4}.
\begin{table}[h]
    \centering
    \caption{Hyperparameters for SVDE.}
    \begin{tabular}{c|c|c}
        \hline
         Hyperparamete & Value & Description \\
        \hline
        Learning rate & $5e-4$ & Step size in each iteration\\
        $\lambda_1$ & $0.99$ & Used for extrinsic reward\\
        $\lambda_2$ & $0.95$ & Used for intrinsic reward\\
        $c$ & $-1e-4$ & Used for explorative experience replay\\
        Gaussian update & $50$ & How often to update $obs$ and $rw$\\
        $\alpha$ & $0.5$ & Weighing $priority$ and $importance\,factor$\\
        $\beta$ & $0.5$ & Weighing $Q^{jt}$ and $Q_{inc}^{jt}$\\
        $\beta_{dec}$ & $1e-4$ & $\beta$ decrements every 1000 train times\\
        Worker number & $3$ & The number of workers we set\\
        Actor number & $4$ & The number of actors each worker set\\
        \hline
    \end{tabular}
    \label{tab:4}
\end{table}

\label{hyper}

\vfill

\end{document}